\documentclass{article}

\usepackage{microtype}
\usepackage{graphicx}
\usepackage{subfigure}
\usepackage{booktabs}
\usepackage{tabularx}
\usepackage{hyperref}
\usepackage{url}
\usepackage{lineno}
\usepackage{graphicx}
\usepackage{algorithm}
\usepackage{algorithmic}
\usepackage{amsmath}
\usepackage{multirow}
\usepackage{booktabs}
\usepackage{amssymb}
\usepackage{hyperref}
\usepackage{caption}

\usepackage[accepted]{icml2025}

\usepackage{amsmath}
\usepackage{amssymb}
\usepackage{mathtools}
\usepackage{amsthm}

\usepackage[capitalize,noabbrev]{cleveref}

\theoremstyle{plain}

\theoremstyle{definition}

\theoremstyle{remark}

\usepackage[textsize=tiny]{todonotes}

\icmltitlerunning{Resolving Lexical Bias in Model Editing}

\begin{document}

\twocolumn[
\icmltitle{Resolving Lexical Bias in Model Editing}



\icmlsetsymbol{equal}{*}

\begin{icmlauthorlist}
\icmlauthor{Hammad Rizwan}{dal}
\icmlauthor{Domenic Rosati}{dal}
\icmlauthor{Ga Wu}{dal}
\icmlauthor{Hassan Sajjad}{dal}

\end{icmlauthorlist}

% \author{Hammad Rizwan, Domenic Rosati, Ga Wu \& Hassan Sajjad\\ 
% % \thanks{ Use footnote for providing further information
% % about author (webpage, alternative address)---\emph{not} for acknowledging
% % funding agencies.  Funding acknowledgements go at the end of the paper.} \\
% Department of Computer Science\\
% Dalhousie University\\
% Halifax, Canada \\
% \texttt{\{hammad.rizwan,domenic.rosati,gw249481,hsajjad\}@dalhousie.edu} \\
% }

\icmlaffiliation{dal}{Department of Computer Science, Dalhousie University, Halifax, Canada}

\icmlcorrespondingauthor{Hammad Rizwan}{hammad.rizwan@dal.ca}

\icmlkeywords{Model Editing, Large Language Models, Continual Learning, Lexical Bias, Representation Learning}

\vskip 0.3in
]



\printAffiliationsAndNotice{\hspace{-1em}} % otherwise use the standard text.

% OLD ABSTRACT
% \begin{abstract}
% Weight-preserving large language model editing techniques rely heavily on scoping mechanisms that determine when to apply edits to the base model. These mechanisms typically use distance functions in the representation space. However, we demonstrate that distance-based scoping functions struggle with strong lexical biases, leading to issues such as applying edits to irrelevant prompts with overlapping words. This paper presents Projector Editor Networks for Model Editing (PENME), a principled approach that learns the optimal representation space for scoping using contrastive learning. Specifically, PENME forms a disentangled representation space that facilitates precise localization of edits by maintaining substantial distance between irrelevant prompts while preserving proximity among paraphrases. In our empirical study, we show PENME achieves state-of-the-art model editing results while being more computationally efficient during inference and adaptable across different architectures.
% \end{abstract}

% DOMs: suggestion
\begin{abstract}
Model editing aims to modify the outputs of large language models after they are trained. Previous approaches have often involved direct alterations to model weights, which can result in model degradation. Recent techniques avoid making modifications to the model's weights by using an adapter that applies edits to the model when triggered by semantic similarity in the representation space. We demonstrate that current adapter methods are \textit{critically vulnerable} to strong lexical biases, leading to issues such as applying edits to irrelevant prompts with overlapping words. This paper presents a principled approach to learning a disentangled representation space that facilitates precise localization of edits by maintaining distance between irrelevant prompts while preserving proximity among paraphrases. In our empirical study, we show that our method (Projector Editor Networks for Model Editing - PENME) achieves state-of-the-art model editing results while being more computationally efficient during inference than previous methods and adaptable across different architectures.
\end{abstract}

\section{Introduction}

\begin{figure}[t!]
   
        \includegraphics[width=0.48\textwidth]{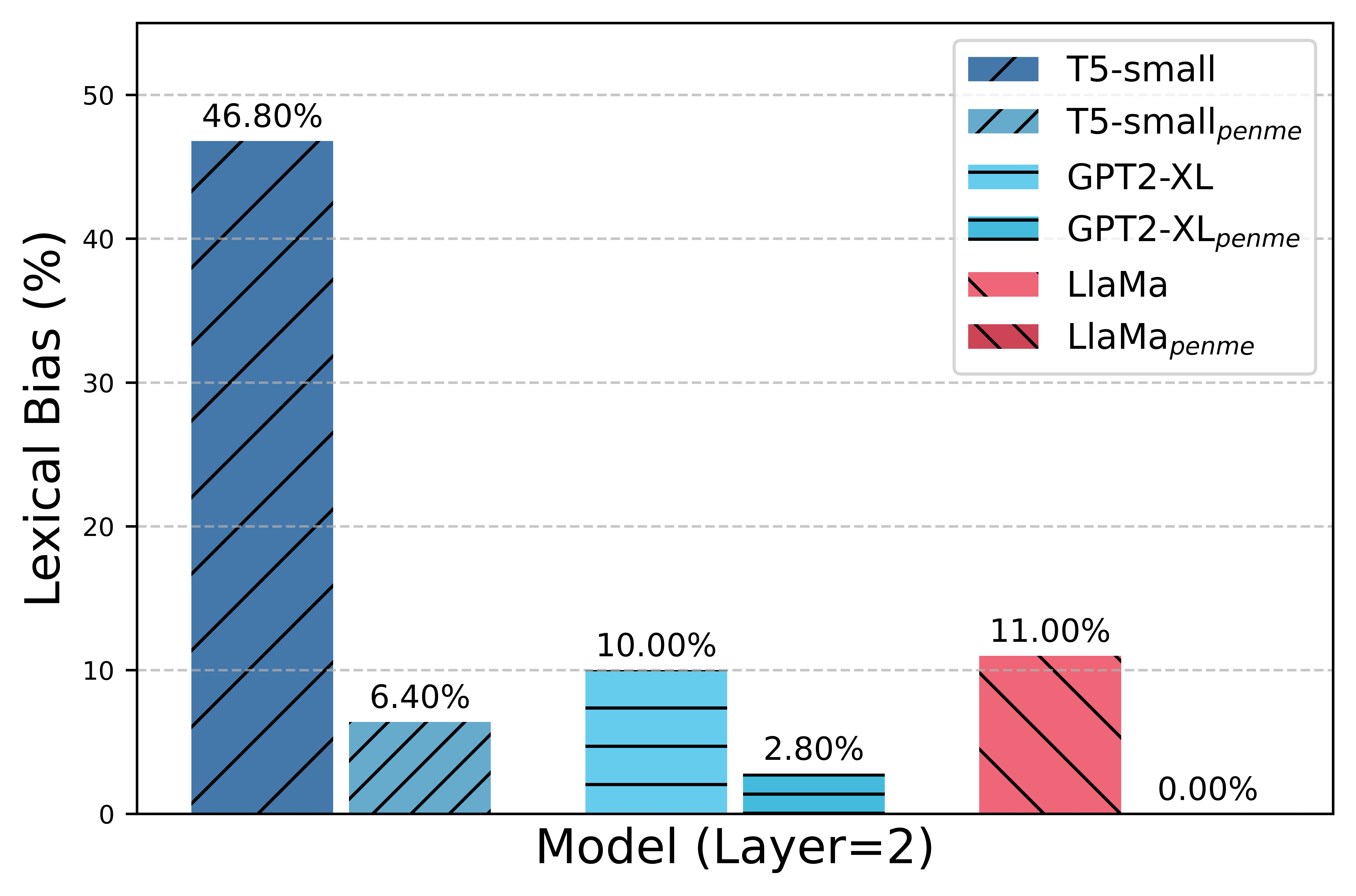}
        \caption{Projector networks mitigate lexical bias: a critical problem in adapter-based model editing techniques. Percentage of samples where irrelevant but lexically similar prompts \textbf{\textit{are closer}} than semantically similar paraphrases in the representation space before and after our learned projection (PENME).}
        \label{fig:lexical_dominance_projector}
        % \vspace{-0.4cm}
\end{figure}

Large Language Models (LLMs) are successful in solving a diverse range of natural language processing tasks \citep{devlin2018bert,liu2019roberta,touvron2023Llama,radford2019language}. Despite their successes, LLMs are fallible in large part due to the noisy and imperfect nature of the data used for training \citep{zhu2020modifying}. As the world evolves, new information requires updates to the models e.g. the leader of a country may change over time.

Periodically retraining LLMs is one potential solution, but risks degrading performance, often requiring training from scratch to maintain previous capabilities \citep{luo2023empirical, wang2023knowledge}. Retraining requires significant computational resources, 
%investment of 
time, data, and skilled labor. 

\begin{figure*}[t]
    \centering
    \includegraphics[width=1\textwidth]{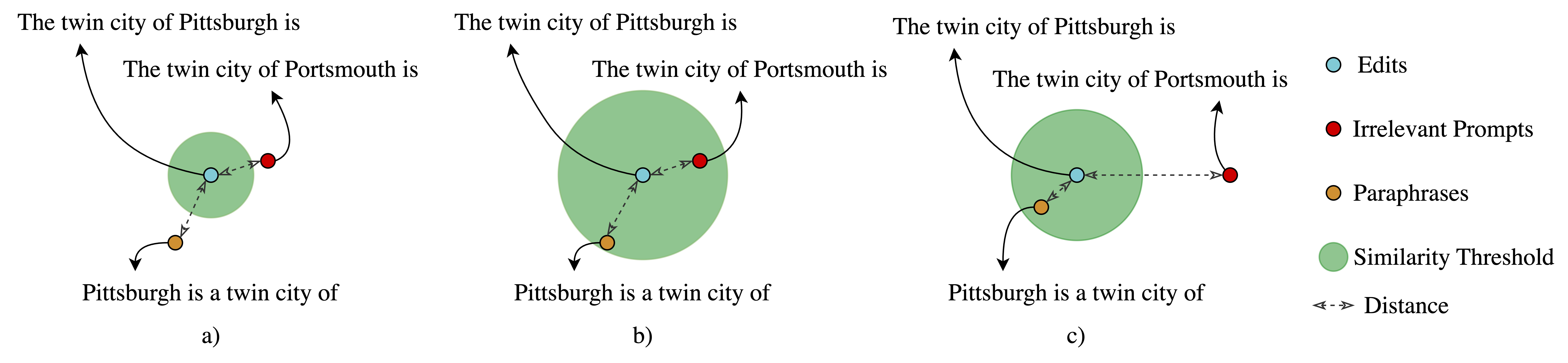} % Change "paper_image1.png" to the filename of your image
    \vspace{-0.6cm}
    \caption{An illustration of lexical dominance in embeddings: a) a low similarity threshold (illustrated with the circle) results in failing to edit paraphrases. b) A similarity threshold results in misfires with irrelevant prompts. c) illustrates our solution which disentangles the representation space.
    }
    \label{fig:similarity_issue}
    % \vspace{-0.4cm}
\end{figure*}

Model editing has been proposed as a sample and compute efficient way to update LLMs \citep{yao2023editing}. Historically \textbf{\textit{weight-modifying}} techniques that make surgical small parameters updates such as ROME \citep{meng2022locating} have been popular. However, these techniques are known to be computationally inefficient \citep{yu2024melo} and result in catastrophic forgetting \citep{gupta2024model}, the full impact of which is difficult to %fully 
determine \citep{rosati-etal-2024-long}.

\textbf{\textit{Weight-preserving }} approaches solve these problems by maintaining the original model parameters and use additional components like key-value codebook adapters (GRACE and MELO) \citep{hartvigsen2024aging,yu2024melo} that look up whether or not to apply a model edit using the semantic similarity between an input and an edit registered in the codebook.

We find that these methods suffer from a critical vulnerability (Figure~\ref{fig:lexical_dominance_projector} and \ref{fig:similarity_issue}). \textbf{Lexical bias} (\textit{prompts with similar lexical tokens but different semantics that are closer together in the representation space as compared to a prompt and its respective paraphrases}) in the representation space prevents current adapter-based methods from effectively being able to balance generalization to unseen semantically similar paraphrases and ``misfiring'' on semantically dissimilar (irrelevant) prompts. Our analysis (\S~\ref{sec:lexical-dominance}; Figure~\ref{fig:lexical_dominance}) of these misfires reveals that the representation space used to calculate semantic similarity is dominated by a lexical bias where irrelevant prompts (e.g. ``The twin city of Portsmouth is") are often closer by euclidean distance to an input (e.g. ``The twin city of Pittsburgh is") than a true semantically similar prompt (e.g. ``Pittsburgh is a twin city of").

% Our work uniquely characterizes this problem of lexical dominance (\S\ref{sec:lexical-dominance}) in relation to model editing and demonstrates that 
% %leading to our novel finding that 
% %for model editing, 
% lexical factors predominantly shape model representations. Figure~\ref{fig:lexical_dominance_complete} shows that 58\% of edits from the Counterfact dataset \citep{meng2022locating} were closer to irrelevant prompts than relevant paraphrase prompts of the original edit prompt using representational similarity measures. This discrepancy leads to misfires in the scoping mechanism, where edits are inappropriately applied to irrelevant prompts. Consequently, this phenomenon creates a trade-off between effectively executing the correct editing mechanism on paraphrases and preventing misfires on irrelevant prompts. This trade-off arises from the suboptimal similarity threshold used to define the boundary between paraphrases and irrelevant prompts. Specifically, a low distance threshold, which constrains the scoping mechanism, minimizes misfires but impedes the successful application of edits on unseen paraphrases. In contrast, a higher threshold enhances the effectiveness of edits on paraphrases but elevates the risk of misfires, as demonstrated in Figure~\ref{fig:similarity_issue}.

% Specifically, a low distance threshold, which controls the scoping mechanism, reduces misfires but impedes paraphrase execution, while a higher threshold enhances paraphrase performance but increases misfire risk as illustrated in Figure \ref{fig:similarity_issue}.

Based on this analysis, we propose to resolve this issue with  Projector Editor Networks for Model Editing (PENME), an advancement over previous adapter-based model editing that explicitly targets the lexical bias problem by learning a projection that disentangles lexically similar and semantically similar text representations. 

% The projector network
% %is a compact, two-layer neural network trained independently 
% uses contrastive learning to disentangle projection space such that paraphrases of edits are close, while irrelevant prompts, both with and without similar lexical overlaps, are farther away. Based on the outputs of the projector network, a memory-based retrieval system facilitates efficient edit retrieval. 
%This approach 
% effectively addresses the aforementioned challenges, while maintaining computational efficiency and ensuring compatibility with both encoder- and decoder-based architectures.

Our contributions are as follows: \textbf{(1)} We demonstrate that representations extracted across layers
%from various LLMs 
exhibit lexical dominance, showing a bias towards token overlap, which introduces significant challenges for adapter-based model editing techniques. \textbf{(2)} We propose PENME\footnote{\url{https://github.com/hammadrizwan/PENME.git}}, a model editing framework that learns a projection network that maps the model's representation space to a new representation space where lexical dominance is minimized. \textbf{(3)} We integrate our projection network in an adapter-based retrieval scheme for model editing, demonstrating, for the first time in adapter-based approaches, high efficacy in both
%both 
paraphrase execution (generalization) and prevention of misfires on irrelevant prompts (locality).
% The proposed projection network is a novel solution to the problem in hand. Moreover, It has broader impact to other application areas relying on representation similarities such as retrieval augmented generation (RAG). However, these applications are out of the scope of this paper.

\section{Related Work}

% Model editing techniques both weight preserving and weight modifying have their associated trade-offs which encompass various factors, including computational requirements for training, inference speed, the need for additional model components or parameters, and overall accuracy. 
% \subsection{Weights modifying}
\textbf{Weight-modifiying}  approaches typically rely on the localization hypothesis (\citealp{miller2016key,geva2020transformer}) in the transformer architecture, which conjectures pointwise feed-forward components function similarly to a key-value memory for information retention within an LLM (a hypothesis which has recently been criticised w.r.t model editing in \citealp{hase2023}). \citet[ROME, MEMIT]{meng2022locating,meng2022mass} identifies salient neurons within the feed-forward layers, facilitating targeted updates to effect the desired edits using causal analysis. Similarly, \citet[PMET]{li2024pmet} investigates the role of multi-headed attention, in conjunction with feed-forward layers, for model editing. As we mentioned earlier, these methods suffer from general model degradation due to gradual performance drift, which can lead to catastrophic forgetting \citep{gupta2024model}. Improvements on MEMIT have been made by \citet[PRUNE]{MaWXLG25}, which bounds the condition number of the edited weight slice so that successive edits cause only minimal drift and the model’s overall behaviour stays intact. Similarly, \citep{FangJWMSW0C25} propose AlphaEdit, which projects each update into the null space of knowledge that must remain unchanged; this preserves existing knowledge while inserting the update. AlphaEdit showcases the strongest performance; however, the method is evaluated on mini-batches of 100 edits, reducing the batch size to $1$ for sequential editing necessitates recomputing the projection for every edit, increasing computational cost and is likely to lead to performance drift during prolonged editing.

An alternative approach, using a hypernetwork, is \citet[MEND]{mitchell2021fast}, which predicts new model weights by generating low-rank decompositions of the weight matrices across different layers.

% Pre-model input approaches necessitate substantial computational resources for two primary purposes: (1) determining relevant contexts and prompts, and (2) processing these contexts and prompts. 
\textbf{Weight-preserving pre-input} approaches depend on extracting and processing relevant edit information before the input is processed by the main model. For example, SERAC~\citep{mitchell2022memory} employs a memory-based model editing strategy augmenting with a memory storage and supplementary models to determine the scope of the edit. 
%Retrieval-augmented (
RAG-based
%)
methods like IKE~\citep{zheng2023can} leverage similarity-based retrieval to extract and rank edit demonstrations from memory and use in-context reasoning to edit. While these are promising approaches, they are generally not computationally efficient as they require additional models for ranking, relevancy, context processing and generalisation.

\textbf{Weight-preserving post-input} These rely on the model's internal representations to implement \textit{scoping mechanisms}: mechanisms that are used to determine whether a specific edit applies for the current input. If an edit does apply, they employ a playback mechanism such as representation vector addition or replacement that results in the model generating updated outputs. Most methods employ a key-value codebook: a vector representation of the inputs which should be edited are stored as a key; and the representation vectors of the desired edit phrase are stored as a value. Semantic similarity is computed using some distance metric, such as euclidean distance, against future inputs to the language model. If a distance threshold is satisfied, then the edit vector is "played-back."

Alternative strategies introduce lightweight auxiliary structures, either additional neurons to scope and steer model outputs, as in \citet{huang2023transformer,zhu2024initializing}, or fixed-size hook layers that accumulate residual updates for consecutive batch edits while keeping the base weights frozen \citet{li2024comebahk}.

\textbf{Lexical Bias} \citet[GRACE]{hartvigsen2024aging} employs playback vectors as above, whereas \citet[MELO]{yu2024melo} utilizes LoRA blocks. We find that both of these methods are vulnerable to lexical bias which invalidates their retrieval methods. In order to resolve this issue, current methods need to hand engineered distance thresholds which balance the trade off generalization for irrelevant prompt misfiring protection. Our method, PENME, is resolves lexical bias by learning a new disentangled representation space where large thresholds can safely be used that are much more reliable at preventing misfires. While lexical bias in NLP has been previously acknowledged, especially in relation to dataset bias \citep{zhou-bansal-2020-towards}, our work is the first to provide an analysis of how this issue effects model editing.

\section{Problem Setting: Model Editing}
\label{sec:prob_dataset}
The objective of model editing is to alleviate the need for complete retraining when updating learned knowledge. Editing attempts to satisfy the following conditions: (1) sample efficiency: update the model with the fewest number of samples possible, (2) compute efficiency: train a small portion of the model only, (3) minimal impact:  make as small of an impact on unrelated behaviour as possible i.e. prevent misfires on irrelevant prompts in adapter-based methods and (4) ensure generalization: maintain accurate paraphrase behaviour i.e. retrieval of correct edits in adapter-based approaches.

The aim is to modify the behaviour of a  model $M$ on a dataset $D=[d_1,...,d_n]$ where the sample $d_i$ is the tuple $(x_i ,y_i,[p_{i1},...,p_{in}],[p^{\neg}_{i1},...,p^{\neg}_{in}])$. $x_i$ is the edit prompt, $y_i$ is the new output tokens, $p_{i,1:n}$ are a set of paraphrases of the edit prompt $x_i$, $p^{\neg}_{i,1:n}$ are 
%\textit{neighbours} or 
\textit{irrelevant prompts} which are examples are both lexically and semantically related; however, they represent cases where the underlying model's generation output should remain unchanged. For instance, consider the edit ``What is the twin city of Detroit." A lexically similar prompt would be ``What is the twin city of London," whereas a semantically related prompt might be ``For Detroit, tell me what twin city it has" where the semantic relationship lies in the fact it is a paraphrase. For successful model editing, the edited model, $M^{\prime}$,  should generate new target tokens $y_i$ for a specific input $x_i$  (Edit Success) and its related paraphrases $p_{1:n}$  (Generalization), while maintaining the model's behaviour on semantically unrelated prompts $p^{\neg}_{1:n}$ (Locality). 
% Typically, datasets for model editing include multiple paraphrases for each edit to assess generalization, along with several instances of , known as neighbors, to evaluate the locality of the edits.
% For any $Dataset=[d_1,d_2,d_3...d_n]$ where the sample $d_i$ is the tuple $(x_i ,y_i,[p_{1i},p_{2i}...],[n_{1i},n_{1i},..])$ where $x$ is the input prompt, $y$ is the new ouput/information, $p$ are the paraphrase prompts, $n$ are the nieghtbourhood prompts.
The following metrics illustrate how these factors are typically operationalized (see for example \citealp{yao2023editing,yu2024melo,hartvigsen2024aging,gupta2024unified}).
% \sout{(model is denoted by $M$, edited model denoted by $M^{\prime}$)}

\begin{figure*}[t]
    \centering
    \includegraphics[width=\textwidth]{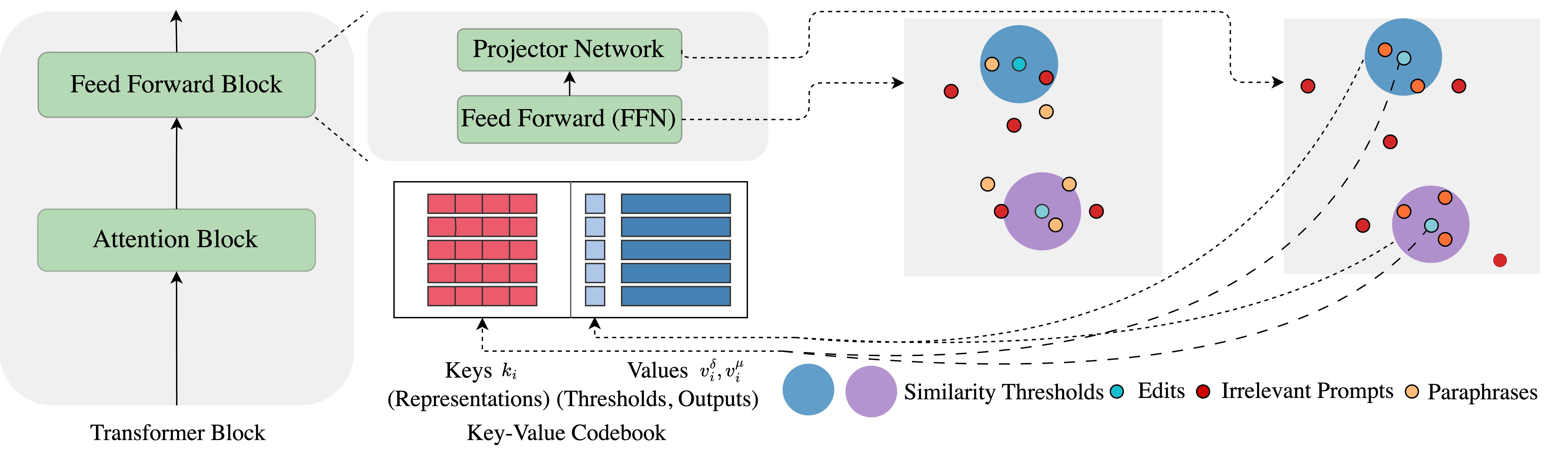} % Change "example-image" to the filename of your image
    \caption{\textbf{PENME} uses a projection network that interfaces with the pointwise feed-forward layer output in a selected transformer block. This projection network, coupled with key-value codebook storage, acts as a scoping mechanism by comparing projection outputs with codebook entries. This mechanism determines whether the current input relates to a specific edit or should pass through the model unmodified.
    }
    \label{fig:PENME}
    % \vspace{-0.4cm}
\end{figure*}

\textbf{Edit Success (ES):} The proportions of edits that the model is able to recall or generate correctly, 
%. This metric has 
also 
referred as
%been called 
efficacy, reliability, and edit score. Formally we say \(
    M^{\prime}(x_i) = y_i, \, \forall (x_i, y_i) \in d_{1:n}\).

    \textbf{Locality:} The proportion of irrelevant prompts for which the model generates the same outputs prior to editing, also referred to as 
    specificity, neighbourhood success, retain rate and neighbourhood score and is denoted as:  
\newline
\(
    M^{\prime}(p^{\neg}_{ij}) = M(p^{\neg}_{ij}), \,\forall p^{\neg}_ij \in d_{1:n}.\)
    
    \textbf{Generalization:} The proportion of paraphrases for which the model is able to recall or generate the correct edited information, also described as paraphrase success: %, paraphrase score:
    \(
    M^{\prime}(p_{ij}) = y_i, \, \forall p_j \in p_i, \forall (p_i, y_i) \in d_{1:n}.
   \)
    
    \textbf{Score:} It 
    %The general score 
    is the mean of the above three metrics and is used for benchmarking.

\section{Projector Editor Networks for Model Editing (PENME)}
\label{sec:penme_overview}
PENME is a weight-preserving model editor that leverages an adapter module which is integrated after the pointwise feed-forward layers within a transformer block of a pre-trained Large Language Model (LLM). By introducing this additional component rather than altering the original model weights, PENME enables the integration of new information while preserving the LLM's initial capabilities.

PENME, illustrated in Figure \ref{fig:PENME}, consists of two components: (1) \textbf{Projection Network ($g$)}
%: this component 
projects model activations denoted $h_l(input)$ at layer $l$ into a distinct representation space $g(h_l(input))$.  
% rojection network that for a specific input projects model activations at layer $l$ to distinct representation space and based on this projection uses a deferall mechanism that decides whether PENME should be used further or default model behaviour is to be used.
(2) \textbf{Key-Value Codebook} 
%that 
stores the projected model activations $g(h_l(input))$ at layer $l$  as keys and corresponding values containing a learned similarity threshold ($\delta$) and the new associated output information $y_i$. This paper only considers storing strings as $y_i$, but vectors \citep{hartvigsen2024aging} or LoRA block indices \cite{yu2024melo} can also be stored as values, which facilitate playback approaches.

In the following sections, the vectors $\vec{x_{ij}}$, $\vec{p_{ij}}$, and $\vec{p^{\neg}_{ij}}$ are outputs of the projection network $g(h_l(input))$ for dataset components $x_i$,$p_{ij}$ and $p^{\neg}_{ij}$

\subsection{Projection Network}
We hypothesize that if the representation space suffers from lexical bias, then we could learn a new representation space that \textit{disentangles} lexical and semantic representations. We achieved this by training a projection network's $g(\cdot): \mathbb{R}^d \to \mathbb{R}^d$ using contrastive learning whose function is to project inputs into a space where paraphrases of inputs are closer to edited inputs than irrelevant prompts. Our training loss is inspired by contrastive learning \citep{hadsell2006dimensionality} and is defined by the following loss function:
%\vspace{-0.3cm}
% \begin{equation}
% \label{eq:example}
% \begin{aligned}
% \mathcal{L}(t,\vec{x_i},\vec{p_{ij}},\vec{n_{ij}}) =  (1-t) \frac{1}{2} ||\vec{x_i} -\vec{p_{ij}}||^2_2 \\
%  + t \frac{1}{2} (\max(0, m - ||\vec{x_i} - \vec{n_{ij}}||_2)^2
% \end{aligned}
% \end{equation}
\begin{equation}
\label{eq:example}
\begin{aligned}
\mathcal{L}(\vec{x_i}, \vec{z}) &= (1-t) \frac{1}{2} ||\vec{x_i} - \vec{z}||_2^2 \\
&\quad + t \frac{1}{2} \big[\max(0, m - ||\vec{x_i} - \vec{z}||_2)\big]^2, \\
t &=
\begin{cases}
1, & \text{if } \vec{z} \gets \vec{p_{ij}}, \\
0, & \text{if } \vec{z} \gets \vec{p^{\neg}_{ij}} \lor \vec{x_l}.
\end{cases}
\end{aligned}
\end{equation}

where $t$ is the target $\{0,1\}$ which is 0 when the training pair is $\{x_i,p_{ij}\}$ (edit, paraphrase) and 1 when the training pair is $\{x_i,p^{\neg}_{ij}\}$ (edit, irrelevant) or the inter-edit (or edit-to-edit) pair $\{x_i,x_l\}$ where we sample an unrelated edit, $m$ is the margin which pushes $\vec{p^{\neg}_{ij}}$ at least $m$ distance away from $\vec{x_{i}}$. The projection network is trained such that for all samples in a dataset, edits $x_i$ and edit paraphrases $p_{ij}$ are close together while edits $x_i$ and irrelevant $p^{\neg}_{ij}$ paraphrases or unrelated edits  $x_l$  are pushed apart in the projection space. Training is performed by sampling pairs at random. Note that $\vec{z}$ is a variable that is assigned either a paraphrase, an irrelevant prompt, or an unrelated edit just as a way to make the loss function more concise.

The inherent lexical and semantic similarities among edits increase the probability of certain edit paraphrases exhibiting greater proximity to other unrelated edits. This phenomenon can lead to erroneous paraphrase-edit associations during execution, potentially triggering inappropriate edit operations. This is why we also push unrelated edits farther away in Eq.~\ref{eq:example} as well as unrelated prompts. These pairings are formed based on a similarity threshold defined as a hyperparameter $\phi$.

The projector network is a 2-layer MLP with one ReLU non-linearity and batch norm applied between each layer. The dimensionality of each layer is the same as the original representation space. Note that this network is only applied to one single layer. The compact architecture of the projection network enables it to be trained on GPUs with limited memory capacity since we can amortize the computation of representation vectors (denoted above with the $\vec{vec}$ symbol) beforehand) irrespective of the underlying model's scale. We provide the details of implementation, data construction and training in Appendix \ref{sec:penme-details}.

\subsection{Key-Value Codebook}
The key-value codebook is a memory mechanism designed to store edits and their corresponding outputs. For each edit, representations are generated by passing the input $x_i$ through the model and the projection network,
%. This is 
denoted as $\vec{x_{i}} =g(h_l(x_i))$. %These representations 
$\vec{x_{i}}$ are then stored as keys $k_i \in K$ in the codebook and are utilized during runtime in a similarity-based retrieval system to access the relevant edit. The codebook value $v_i \in V$ consists of the edited information $\mu$ along with a similarity threshold $\delta$. The edited information in this paper is an exact string stored from the model editing dataset. The threshold serves as a scoping mechanism and is learned using a procedure described below. For a given input prompt 
%denoted 
$x_i$, euclidean distance $||\cdot||_2$ is computed with all keys in the codebook. From the computed distances, we determine if the input prompt $x_i$ is relevant to the edited codebook value $v^{\mu}_i$ and its corresponding threshold $v^{\delta}_i$. This is expressed as:
% \begin{equation}
% \label{eq:distance_threshold}
% \begin{alignedat}{2}
% &\underset{k_i}{\arg \min}\:\: ||\vec{x_{i}} - k_i||_2 \\
% & \text{s.t.  } ||\vec{x_{i}} - k_i||_2 < v^{\delta}_i
% \end{alignedat}
% \end{equation}

\begin{equation}
\label{eq:distance_threshold}
\begin{alignedat}{2}
&\underset{k_i \in K}{\arg\min}\:\: \left\|\vec{x_i} - k_i\right\|_2 \\
& \text{s.t.} \quad \left\|\vec{x_i} - k_i\right\|_2 < v^{\delta}_i
\end{alignedat}
\end{equation}

If the prompt $x_i$ is deemed relevant 
%based on the 
(Equation~\ref{eq:distance_threshold}), the output information of the edit is retrieved from codebook $v^{\mu}_i$. Otherwise, the typical model output $M(x_i)$ is employed.

\subsection{Finding the thresholds $v_i^{\delta}$ and $\tau$}
Initial experimental findings regarding the thresholds $v_i^{\delta}$ reveal that unseen test paraphrases typically demonstrate greater distance than the average seen training paraphrases, while the inter-paraphrase distances within the training set exhibit variation across edits. In contrast, unseen test irrelevant prompts generally show closer proximity to edits compared to the nearest seen training irrelevant prompts. This effect is illustrated in greater detail in Appendix \ref{sec:pre_post_proj}.  
%To determine an 
We determine an appropriate threshold 
%that defines the scope of an edit, we 
by utilizing a data-driven thresholding scheme based on the training data:

\begin{equation}
    v^{\delta}_i =\text{Max} \left( \lVert \vec{x}_i - \vec{p_{ij}} \rVert_2 \right) + \tau 
\label{eq:thresholding}
\end{equation}

% \begin{equation}
% \begin{alignedat}{2}
% Max(||\vec{x} - \vec{p_{ij}}||_2) + \tau \\ 
% \text{setting} \tau \text{distance away from max paraphrase distance}
%     \label{eq:thresholding}
% \end{alignedat}
% \end{equation}
% . Based on this analysis, we conclude that using option 1. is not suitable.  Option 2 or 3, utilizing maximum and minimum distances, appears to be the most appropriate choice. 

% The decision between the two depends on the perspective from which adjustments need to be made. Option 2 ensures locality with all training neighbors maintained, whereas option 1 provides assurance that all training paraphrases will function irrespective of the chosen alpha. For our experiments, we use option 1 as it guarantees a full edit success rate.
The threshold is determined as the maximum paraphrase distance observed for each individual edit, augmented by a hyperparameter $\tau$ to account for unseen paraphrases. The hyperparameter $\tau$ is found through grid search. This formulation allows our method to achieve an optimal balance between generalisation and locality preservation. Alternatively, another possibility is to set the threshold based on close irrelevant prompts $Min(||\vec{x} - \vec{p^{\neg}_{ij}}||_2) - \tau$, this option would maintain locality by preserving all training irrelevant prompts.

\subsection{Analysis of Codebook Management and Scalability} Both GRACE and MELO require multiple paraphrases added to the codebook to improve generalisation. The PENME codebook scales linearly with the number of edits, as each edit corresponds to a single codebook entry. Maintaining one entry per edit enables efficient edit removal or updates, providing greater flexibility in edit management. In contrast, the scoping mechanism employed by \citet{hartvigsen2024aging,yu2024melo}  to deal with multiple, possibly conflicting, entries per codebook requires splitting and merging operations. The effectiveness of this approach varies across datasets. For instance, for GRACE, the zsRE dataset exhibits a high occurrence of similar edit outputs (same entity with the same edit), allowing for substantial reductions in codebook entries. Specifically, 1,000 edits on zsRE require only 658 entries, whereas the Counterfact dataset requires 1,682 entries for just 300 edits. The combination of this consolidation process and the potential for edits to be closely related in vector space leads to overlapping cluster radii, necessitating cluster size reduction. This inadvertently results in the removal of certain edits. A detailed comparison between PENME and the scoping mechanisms employed by GRACE and MELO is presented in Appendix \ref{sec:penme_vs_grace_melo_scope}. The results demonstrate that PENME achieves superior edit retrieval speed and highlights the problem of edit conflict and edit forgetting by GRACE and MELO.

\section{Experimental Setup}
% We select the datasets, baselines, and models that are most common in the model editing literature.

We assess the performance of PENME across a spectrum of transformer-based LLMs including %These models include 
Text-to-Text Transfer Transformer (specifically T5-small) \citep{raffel2020exploring}, Llama-2-7b~\citep{Llama} and GPT2-XL \citep{radford2019language}. 
%We compare PENME with 
We compare PENME with GRACE and MELO, as these are the only other current weight-preserving adapter-based methods (\citet{wang2024wise} is contemporaneous work that was published after the writing of this paper). Additionally, we include MEMIT and SERAC\footnote{A simpler version of SERAC is used in \citet{hartvigsen2024aging} called Defer.}.Working details of the methods and hyperparameters are provided in Appendix \ref{sec:experimental-setup}. 
%To select the optimal layer to introduce the PENME adapter, 
We utilize the methodology outlined in \S \ref{sec:lexical-dominance} to select an optimal layer to introduce PENME adapter and use the second layer for all LLMs. We determine the optimal threshold for each edit 
by systematically varying the $\tau$ parameter in Equation \eqref{eq:thresholding} across a range of $0.05$ to $0.20$. 
\label{para:dataset}
\paragraph{Dataset}
The zsRE dataset \citep{levy2017zero} and the Counterfact dataset \citep{meng2022locating} are the most commonly used 
 model editing datasets. zsRE consists of an edit prompt along with several paraphrased versions of that prompt.
 %To evaluate the impact of edits on unrelated knowledge, 
 Irrelevant prompts are sourced from the NQ dataset~\citep{kwiatkowski2019natural}, which offers a wide range of user query questions.
In contrast, Counterfact has similar edit and paraphrase prompts but employs a more nuanced approach to irrelevant prompts. It includes prompts that are similar to the edit prompt in both semantic nature and lexical structure. This differs significantly from zsRE, where the irrelevant prompts are neither semantically nor lexically related to the edit prompt. Moreover, zsRE has a lower diversity in subjects, relationships, and linguistic variations \citep{meng2022locating}.
This structural difference between the datasets has important implications for evaluation. In zsRE, the lack of semantic or lexical relationships between the edit prompt and its irrelevant prompts allows weight-preserving approaches to achieve high locality scores with relative ease. The enhanced complexity of Counterfact renders it a more robust benchmark for evaluating editing mechanisms. Dataset processing and training data construction details are provided in Appendix~\ref{sec:counterfact-data-processing}.

\paragraph{Downstream Tasks} To assess downstream performance, we adopt a similar evaluation setup to that used by \citet{MaWXLG25}. Specifically, we evaluate our approach on three tasks: sentiment classification using the DAiR-Emotions dataset~\citep{saravia-etal-2018-carer}, summarisation using the CNN/DailyMail dataset~\citep{hermann2015teaching}, and natural language inference (NLI) using the RTE dataset~\citep{dagan2005pascal}.

% is designed to evaluate the editing of relation extraction information. Each entry in this 

% Furthermore, the Counterfact dataset has a broader spectrum of subjects, relationships, and linguistic variations

% to test if other random facts ha The samples in these datasets are composed of an edit prompt, accompanied by multiple paraphrases and neighboring prompts.

\section{Evaluation}

This section presents the evidence of lexical bias, the results of PENME in achieving separability of irrelevant prompts and paraphrases, and a comparison with other methods.

\begin{figure}[t!]
        \includegraphics[width=0.48\textwidth]{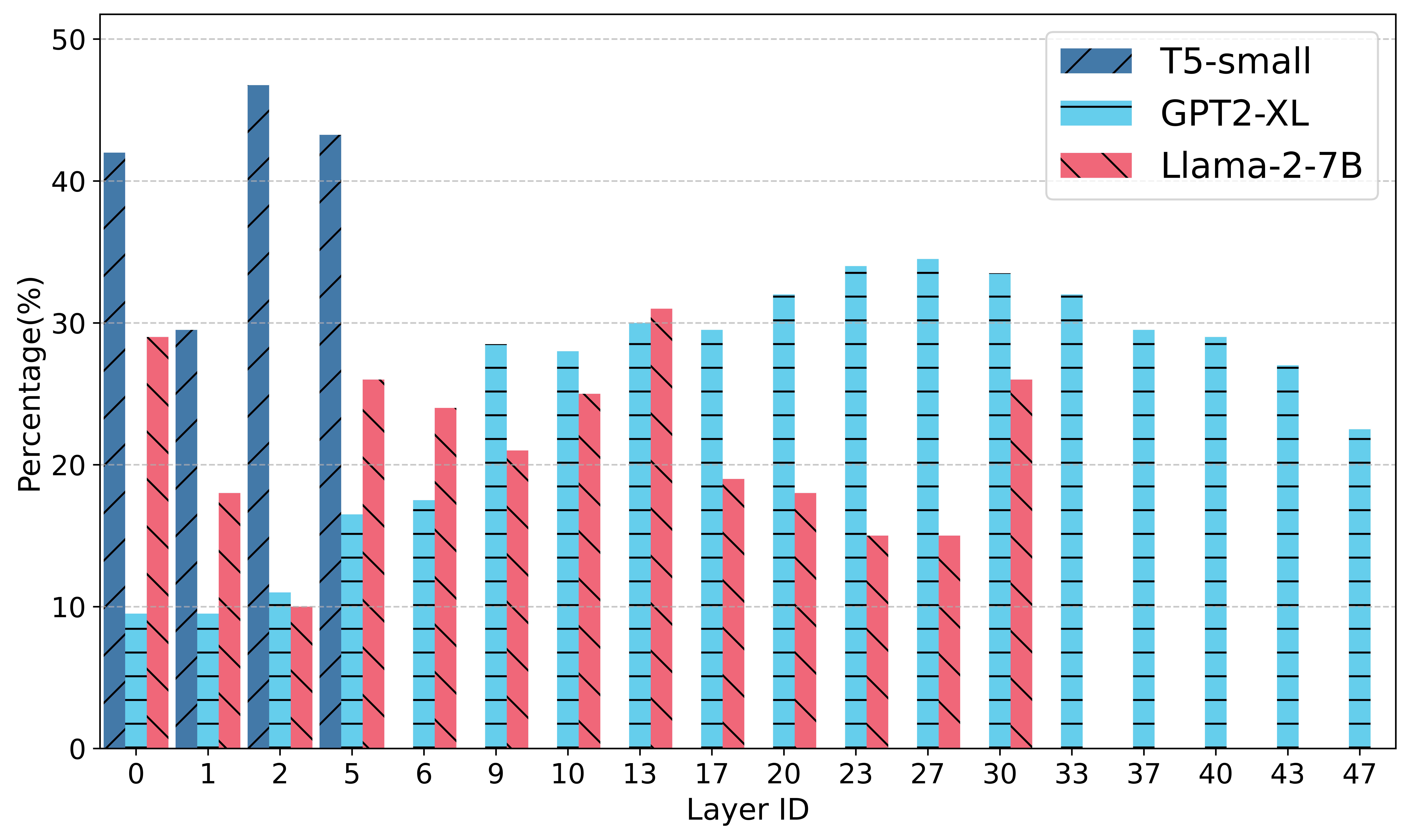}
        \caption{Percentage of samples \textit{where edits are closer} to lexically similar yet irrelevant prompts as compared to paraphrases in the representations space of different models across various 
        %select 
        layers. T5-small, GPT2-XL and Llama-2-7b have 6, 32, and 48 layers, respectively. The full figure for all layers can be found in Appendix \ref{sec:lexical_dominance_details}.}
        \label{fig:lexical_dominance}
        \vspace{-0.4cm}
\end{figure}

\subsection{Lexical Bias}
% in Model Representation
\label{sec:lexical-dominance}

To examine the lexical dominance of representations, 
%in Large Language Models (LLMs),
we randomly sampled 500 entries from the Counterfact dataset (see \S \ref{para:dataset}). For each entry, we created triplets consisting of an edit prompt, a randomly sampled paraphrase prompt and an irrelevant prompt with \textbf{\textit{high lexical overlap}} (\(x_i,p_i,p^{\neg}_i\)). In Table~\ref{tab:token_metrics} we see the ROUGE-1 token overlap w.r.t $x_i$ mean F1 was 0.3 for both $p_i$ and $p^{\neg}_i$. Qualitative samples are provided for the readers validation in Table~\ref{tab:data_samples}. These triplets are fed into various models, and representation vectors ($\vec{x_{i}},\vec{p_{i}},\vec{p^{\neg}_{i}}$) from the feed-forward block of each layer $l$ are extracted. 
%The choice of representations depends on model capabilities. 
We select either averaged token representations or dedicated sentence representations, based on whether a given model offers a specific token for sentence-level representation. Following extraction, we calculate two sets of pairwise Euclidean distances: (1) Between edit representations and paraphrase representations: $||\vec{x_{i}} -\vec{p_{i}}||_2$ (2) Between edit representations and irrelevant prompts representations: $||\vec{x_{i}} -\vec{p^{\neg}_{i}}||_2$. We then compare these distances to determine if irrelevant prompts are closer to the edits than the paraphrases $||\vec{x_{i}} -\vec{p_{i}}||_2 > ||\vec{x_{i}} -\vec{p^{\neg}_{i}}||_2$. Figure \ref{fig:lexical_dominance} displays the percentage of samples where irrelevant prompts \textit{were closer} to the edits.

The findings in Figure~\ref{fig:lexical_dominance} reveal an intriguing pattern: except for the first layer in most models, the early layers demonstrate a reduced percentage of samples where irrelevant prompts are closer to edits than paraphrases. However, the trend shifts as we progress through the model's depth. In the mid-layers, this percentage begins to ascend once more, only to descend slightly towards the final layers, albeit with subtle fluctuations among them. We hypothesize that in the initial layers, token-specific information remains largely isolated. However, as the input traverses deeper into the model, guided by repeated attention mechanisms, this information becomes amalgamated across tokens. Moreover, repeated normalization as demonstrated by \citet{takase2022layer} results in smaller changes in weights of an LLM leading to embedding vectors in the final layers being similar, thus only subtle fluctuations are seen in the percentages. This issue of lexical bias is not limited to model editing but  is also highlighted by \citet{dumpala2024sugarcrepe++} who examine the
impact of lexical diversity on model representations for semantically equivalent texts and showed
that their representations often exhibit divergence despite semantic equivalence

%\hs{following is a sudden jump to identifying optimal layer. One may first reiterate that these results demonstrate the reason behind misfire prompt. They also provide a systematic way to identify an optimal layer for ... by the way do we need to find an optimal layer or our prejection network will work on any layer but the efficacy will change depending on which layers are easier to disentangled}
% \hr{text updated, It takes more time to train as similarities are high to begin with, the only test I did was on T5 for 500 samples other layers are able get somewhat similar results but this needs to be tested on a larger scale (2000 edits minimum)}

These results indicate why there is a significant chance of misfire in adapter-based methods: lexical bias. This also provides a systematic approach for identifying the optimal layer to introduce PENME integration, by elucidating the regions within the model's architecture where lexical dominance exhibits minimal influence. Although the projector network approach can be generalized across all layers, as demonstrated in Appendix~\ref{sec:layer_abalation}, it is advantageous in terms of training time to integrate at points of minimal influence.
% establishes a systematic framework for identifying the optimal layer for adapter integration, by elucidating the regions within the model's architecture where lexical dominance exhibits minimal influence.

\begin{table*}[t]
\caption{A comparative analysis of PENME and recent model editing methods on 2000 edits from the Counterfactual dataset and 1000 edits on zsRE. The metrics are Edit Success (ES), Locality (Loc) and Paraphrase Generalization (Para).}
\scriptsize
    \centering

    \begin{tabularx}{\textwidth}{p{2cm}p{2cm}p{1.15 cm}p{1.15 cm}p{1.15 cm}p{1.15 cm}|p{1.15 cm}p{1.15 cm}p{1.15 cm}p{1.15 cm}
    % p{0.06\textwidth} p{0.18\textwidth} >
    % {\centering\arraybackslash}p{0.03\textwidth} >
    % {\centering\arraybackslash}p{0.03\textwidth} >
    % {\centering\arraybackslash}p{0.03\textwidth} >
    % {\centering\arraybackslash}p{0.03\textwidth} >
    % {\centering\arraybackslash}p{0.03\textwidth} >
    % {\centering\arraybackslash}p{0.03\textwidth} >
    % {\centering\arraybackslash}p{0.03\textwidth}
    }
     
        \toprule
        & & \multicolumn{4}{c}{\textsc{Counterfact}} & \multicolumn{4}{c}{\textsc{zsRE}} \\
        \textbf{Method}  & \textbf{Model} & \textbf{ES} & \textbf{Loc} & \textbf{Para} & \textbf{Score} & \textbf{ES} & \textbf{Loc} & \textbf{Para} & \textbf{Score} \\
        % \midrule
        % Defer 
        % &   T5-small & \textbf{1.0} & 0.136 & 0.271 & \textbf{0.} 
        % & \textbf{1.0} & 0.678 & 0.437  &  \textbf{0.0}  \\
        % &   Llama-2-7b  & \textbf{1.0}& 0.7617 & 0.544  & \textbf{0}  
        % & \textbf{1.0}  & 0.98 & 0.46 & \textbf{0.0}  \\
        % &   GPT2-XL & \textbf{1.0} & 0.799 & 0.670 & \textbf{0.}
        % &\textbf{1.0}  & 0.719 & 0.879 & 0.0 \\
        \midrule
        PENME 
        &   T5-small & \textbf{1.000} & 0.787 & \textbf{0.808} & \textbf{0.865} 
        & \textbf{1.000} & \textbf{0.941} & 0.913  &  \textbf{0.951}  \\
        &   Llama-2-7b  & \textbf{1.000} & 0.869 & \textbf{0.906}  & \textbf{0.925}  
        & \textbf{1.000}  & \textbf{0.987} & \textbf{0.966} & \textbf{0.984}  \\
        &   GPT2-XL & \textbf{1.000} & 0.847 & \textbf{0.875} & \textbf{0.907}
        &\textbf{ 1.000}  & \textbf{0.957} & 0.940 & \textbf{0.966} \\

        \midrule
        MELO 
        & T5-small & 0.850 & 0.800  & 0.037 & 0.562 &  0.990 & 0.640 & \textbf{0.986} &  0.872\\
        & GPT2-XL & \textbf{1.000} & \textbf{1.000}  & 0.020 & 0.673 & \textbf{1.000} & 0.004 & \textbf{1.000} & 0.668 \\

       \midrule

        GRACE  
        & T5-small & \textbf{1.000} & \textbf{0.860} & 0.140 & 0.667 & \textbf{1.000} & 0.730 & \textbf{0.993} & 0.907  \\
        % &  Llama-2-7b & \textbf{1.000} & \textbf{0.997} & 0.002 & 0.666 & 0.120* & 0.000* & 0.579* &  0.233* \\
        % &  GPT2-XL& \textbf{1.000} & 0.996 &  0.003 & 0.666 & 0.993* & 0.019* & 0.017*  & 0.343*\\
        & Llama-2-7b & \textbf{1.000} & \textbf{0.997} & 0.002 & 0.666 & 0.100 & 0.591 & 0.000 &  0.230 \\
        &  GPT2-XL& \textbf{1.000} & 0.996 &  0.003 & 0.666 & 0.992 & 1.000 & 0.010 & 0.667 \\
   
        \midrule
        SERAC 
        & T5-small & 0.017 & 0.526 & 0.010 & 0.184 & 0.017 & 0.526 & 0.010 & 0.184 \\
        & Llama-2-7b & 0.992 & 0.372 & 0.651 & 0.672 & \textbf{1.000} & 0.114 & 0.357  & 0.490 \\
        & GPT2-XL & 0.947 & 0.669 & 0.408 & 0.675 & 0.474 & 0.003 & 0.811 & 0.429\\
        \midrule
        MEMIT 
        & Llama-2-7b  & 0.147 & 0.149 & \textbf{1.000} & 0.432 & 0.402 & 0.002 & \textbf{1.000}  & 0.468\\
        & GPT2-XL  & 0.785 & 0.788 & 0.502 & 0.692 & 0.214 & 0.000 & \textbf{1.000}  & 0.405\\
        
        \midrule
        FT
        & T5-small & 0.955 & 0.000 & 0.450 & 0.468 & 0.017 & 0.526 & 0.010 & 0.184 \\ 
        & Llama-2-7b & 0.404 & 0.393 & 0.417 & 0.405 & 0.569 & 0.020 & 0.746 & 0.445\\
        & GPT2-XL & 0.968 & 0.851 & 0.395 & 0.738  & 0.608 & 0.005 & 0.889  & 0.501 \\
        % \midrule
        \bottomrule
        $\text{PENME}_{\text{stream}}$ 
        & T5-small  & 1.000    & 0.782    & 0.756    & 0.846 & 1.000    & 0.615    & 0.550    & 0.721    \\ 
        & Llama-2-7b & 1.000 & 0.871    & 0.818    & 0.896 & 1.000     & 0.716    & 0.792    &  0.836      \\ 
        & GPT2-XL  & 1.000    & 0.850    & 0.768    & 0.872  & 1.000    & 0.733    & 0.768    & 0.833 \\

        \bottomrule
        \end{tabularx}
    
    \label{tab:model_editing_experiments}
    \vspace{-0.4cm}
\end{table*}

\subsection{Disentangled Projection Space}

% \hs{this is a good terminology that should have been mentioned in the introduction while providing the intuition of our proposal}\hr{added to introduction}
% \hs{we are using "each of the models" in several places but we never introduce our models till now. When you are mentioning about a method, you can simply assume that there is one given model or call it subject model} 
% \hs{clearly state the data splits. We first split the data and then we train and test ..}
% \hr{paragraph updated}
In this section, we validate our proposed projection network in its ability to learn a generalized disentangled representation space where paraphrases are closer to edits as compared to irrelevant prompts. We sample 1500 tuples (\(e_i,p_i,p^{\neg}_i\)) of edits denoted $e_i$, paraphrases $p_i$, and their unrelated irrelevant prompts $p^{\neg}_i$ from the Counterfact dataset with accompanying input prompts $x_i$ and split them into train and test sets of 1000 and 500 samples respectively. We use the training set to train the projector network using model representations from layer 2 of each model.
%for all models on the prepared data. 
To evaluate the network's performance, we compare two types of test representations: the original model representations $h_l(x_i)$ where $x_i$ is the input prompt and the projected representations $g(h_l(x_i))$. This comparison uses the experimental method described earlier, allowing us to determine whether the projection network successfully learns to create a lexically disentangled representation space.
% After training the projector network, we compare the model representations $h_l(x)$ with those of the projection network $g(h_l(x))$ using the previously described experiment.

The results presented in Figure \ref{fig:lexical_dominance_projector} demonstrate that the projector network effectively learns to distance lexically similar but unrelated irrelevant prompts in comparison to paraphrases.
A two-dimensional PCA visualization of the representation space, illustrating this phenomenon, is provided in Appendix \ref{sec:pca_visualization}.
% \hs{I would suggest to add results of T5 in appendix. You may also select four to five layers from different parts like initial, middle, last and show the robustness of our classifier. You may add reference to the appendix in the experimental setup}

%filters out lexical information from the model representations and focuses on the semantics of the text. 
%
For data pairs where irrelevant prompts are closer to edits than paraphrases, T5-small exhibits a dramatic decrease from $46\%$ to $6.4\%$. Similarly, GPT2-XL reduces from 10\% to 2.8\%, and Llama-2-7b drops to $0\%$ from 11\%, indicating perfect separability of irrelevant prompts and paraphrases.

\subsection{Model Editing Results}

Table \ref{tab:model_editing_experiments} presents the comparative results of PENME and recent model editing methods for 2000 edits on the Counterfact dataset and 1000
% \footnote{*Due to system implementation issues with GRACE on EasyEdit \citep{wang2023easyedit}, we were only able to compute a 100 sample subset for results with a *.} 
edits on zsRE\footnote{Due to computational constraints, editing is performed on 1,000 zsRE samples. However, this number is consistent with the typical sample size used in related literature.}. 
%From the results, we determine that 
PENME demonstrates a highly stable performance across editing metrics as compared to other model editing approaches. In particular, PENME shows high efficacy on both locality and generalization and has stable performance across the different models. Observe that for both GRACE and MELO, these methods require trading of locality for paraphrase performance or vice versa due to lexical bias.

GRACE, similar to PENME, demonstrates high edit success rates due to its inherent design. However, its generalization scores compared to PENME were markedly low, suggesting poor performance on edit paraphrases post-editing. GRACE achieved the highest locality scores, with T5-small at 0.92 and Llama-2-7b nearly perfect at 0.997. The substantial difference between locality and generalization scores can be attributed to GRACE's use of a very low distance threshold, resulting in poor performance on paraphrases but successfully avoiding irrelevant prompts spillover into edits.

SERAC 
%is 
also 
%able to 
achieves
a high edit success but shows mixed performance results for generalization and locality across models. For T5-small, the approach does not work well as SERAC uses logically entailed facts to determine the scope, the original work uses a T5-large which is significantly better at reasoning.

For GPT2-XL, MEMIT demonstrates moderate effectiveness, achieving an edit success rate of 0.785 and a locality score of 0.788. In contrast, when applied to Llama-2-7b, both the edit success and paraphrase success rates are relatively low, although the locality score remains high. This discrepancy is likely due to challenges stemming from MEMIT's training on the Llama-2-7b model, as similar findings have been reported by \citet{huang2024reasons}.

\begin{figure*}[t]
    \centering
    \includegraphics[width=\textwidth]{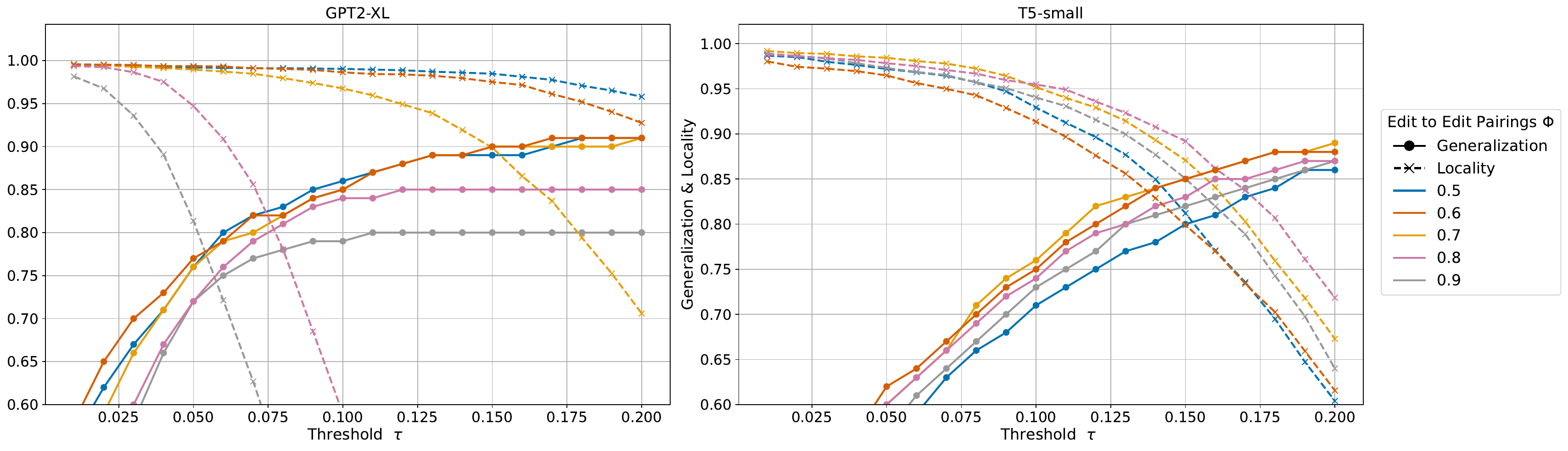} % Change "example-image" to the filename of your image
    \caption{Shows the trade-off between generalization and locality performance across different hyperparameter settings. The distance threshold $\tau$ varies from $0.01$ to $0.2$ ($0.01$ increments and $\tau$ is normalized by 100), while the edit-pairing similarity threshold $\phi$ ranges from $0.5$ to $0.9$ ($0.1$ increments). Higher $\phi$ values enforce stricter edit similarity requirements. The results showcase the effect of hyperparameter tuning on the projector network's learning capacity and overall performance.}
    \label{fig:generalization_vs_locality}
    % \vspace{-0.4cm}
\end{figure*}
\textbf{Projector Generalization} In the previous setting, we trained the projector using all of the edit samples at once i.e. batch editing. In this setting, we evaluate the stream or lifelong editing setting for zero-shot generalization, where we update the codebook once per edit using a frozen projector. As a trained projection network is needed we initialize $\text{PENME}_{\text{stream}}$ using 2k unseen samples from Counterfact.  Results from these experiments are presented in Table \ref{tab:model_editing_experiments} under $\text{PENME}_{\text{stream}}$. Analysis of the ZsRE dataset demonstrates the projector network's capacity for zero-shot generalization, achieving robust performance metrics while maintaining equilibrium between generalization and locality. For the Counterfact dataset, the drop in performance is minimal with the slight exception on the generalization of GPT2-XL. These findings suggest that training a projector network on a more extensive dataset to reduce lexical bias could enable its use as a modular component.

\begin{figure}[t]
    \centering
    \includegraphics[width=0.48\textwidth]{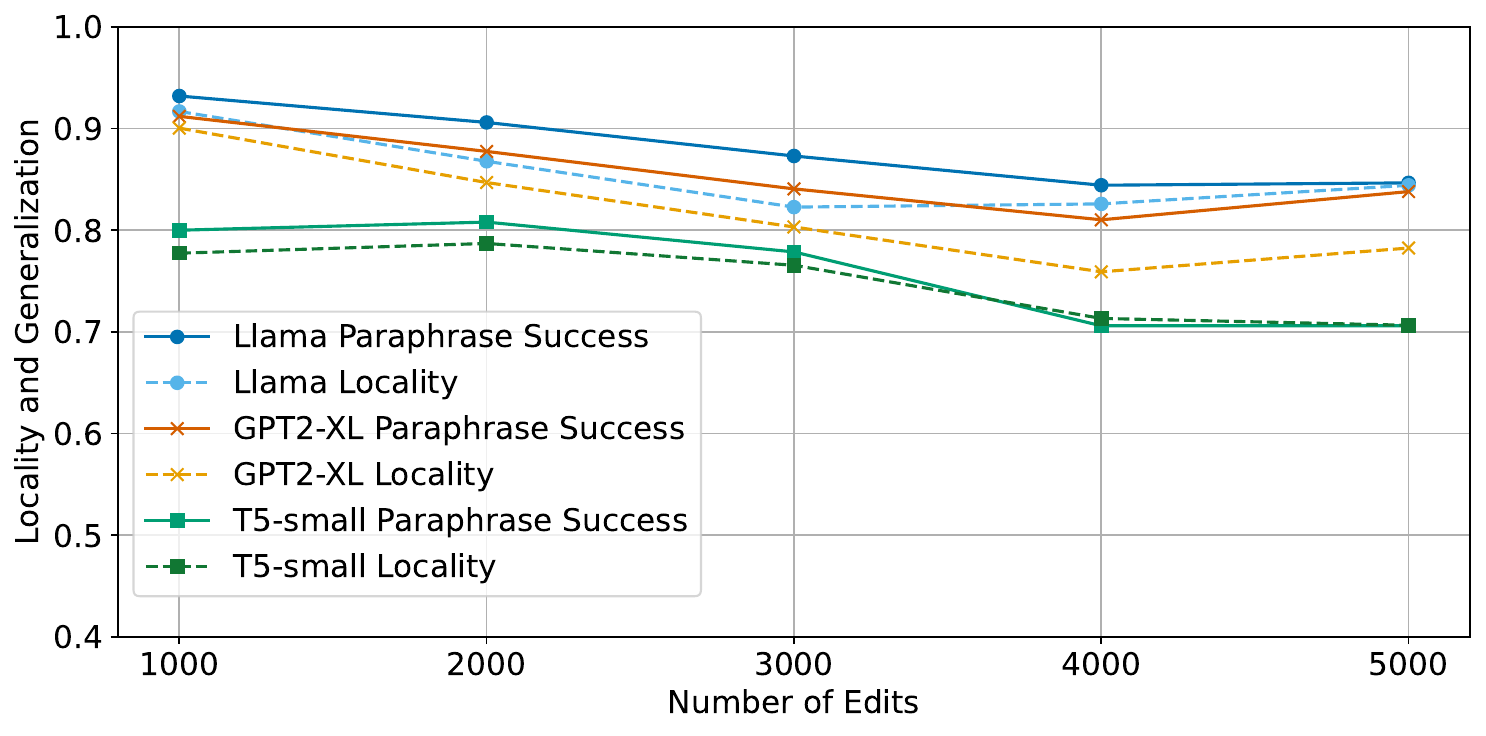} % Change "example-image" to the filename of your image
    \caption{PENME's performance in terms of Locality (dotted) and Generalization (continuous line) across varying numbers of edits}
    \label{fig:scaling_edits}
    % \vspace{-0.4cm}
\end{figure}
\subsection{Scaling Edits}
We evaluate the projection network's stability under varying numbers of edits using incrementally larger training sets ranging from 1000 to 5000 edits, with 1000-edit increments per training session. The results of the experiment are shown in Figure~\ref{fig:scaling_edits}. 
Projector network trained on representations from T5-small demonstrates lower overall performance in generalization and locality compared to other models. We hypothesize that this under-performance may be attributed to either the model's smaller size, resulting in less robust learned representations, or the fact that it was trained on a more limited dataset relative to larger, more recent models. Projection networks trained on Llama-2-7b and GPT2-XL representations exhibit comparable performance levels. Both models show a slight decrease in generalization and locality performance as the number of edits increases from 1000 to 2000, with minimal decline after that. 

Examination of projection network behaviour reveals interesting patterns in generalization and locality failures based on the varying distances between training edits and their respective paraphrases and irrelevant prompts after the training of the projector network. The varying distances result in different thresholds for each edit, which can cause errors when the closest edit to an irrelevant prompt example has a high threshold. To quantify these observations, we employed ROUGE scores in a comparative study of generalization outcomes. Appendix \ref{sec:token_overlap_analysis} provides further analysis of the learned projection space.

\begin{table*}[t]
\centering
\scriptsize
\begin{tabular*}{\textwidth}{@{\extracolsep{\fill}}l c c c}
\toprule
\textbf{Model} & \textbf{NLI} & \textbf{Sentiment Classification} & \textbf{Summarization} \\
\midrule
Llama-2-7b & 0.6476 & 0.6573 & 0.1865 \\
$\text{Llama-2-7b}_{\text{PENME}}$ & 0.6428 & 0.6573 & 0.1865 \\
GPT2-XL & 0.5128 & 0.4630 & 0.0936  \\
$\text{GPT2-XL}_{\text{PENME}}$  & 0.5128  & 0.4630   & 0.0936  \\
\bottomrule
\end{tabular*}
\caption{Downstream task performance of GPT2-XL and Llama-2-7B before and after PENME editing, demonstrating the method's effect on model generalization capabilities.}
\label{tab:nlp_comparison}
\end{table*}

\begin{table}[t]
\centering
\scriptsize
\begin{tabular}{l c c}
\toprule
\textbf{Model} & \textbf{Fluency} & \textbf{Reference Score}  \\
\midrule
Llama-2-7b &  611.54 & 16.57 \\
$\text{Llama-2-7b}_{\text{PENME}}$  & 622.36 & 21.98   \\
\bottomrule
\end{tabular}
\caption{Evaluation of long-form generation by Llama-2-7B pre- and post-editing with PENME (via IKE).}
\label{tab:generation_metrics}
\end{table}
\section{Generalization and Locality}

% These experimental findings suggest that training a projector network on a more extensive dataset to mitigate Lexical Bias effects could facilitate its deployment as a modular component.

 % We conducted experiments using a network trained on the Counterfact dataset in the previous experiment while maintaining consistent hyperparameters. The pretrained projector was subsequently evaluated on 1,000 unseen editing instances drawn from both the ZsRE and Counterfact datasets.
% To demonstrate the projector network's generalizability as a plug-and-play component for model editing, we evaluated its performance using a network pretrained on the Counteract dataset with the same hyperparameters. The pretrained projector was applied to 1,000 \textit{previously unseen} edits across both ZsRE and Counterfact datasets. Table \ref{tab:model_editing_experiments} $PENME_{stream}$ shows the results of the experiment. We observe that for ZsRE that the projector network is able to zero-shot generalize by achieving high scores with a balance between generalization and locality. For Counterfact the model achives 
% \subsection{Generalization and Locality}
To demonstrate the trade-off between generalization and locality, we conducted an ablation study by varying the $\tau$ parameter, which modulates the similarity threshold defining an edit's scope. Figure~\ref{fig:generalization_vs_locality} presents the results for GPT2-XL and T5-small. The trends observed for GPT2-XL and Llama-2-7b are similar. Therefore, for clearer visualization, we present the detailed results for Llama-2-7b  separately in Appendix \ref{sec:gen_vs_loc_models}. Setting a low $\tau$ value achieves near-perfect locality but poor generalization. As we incrementally increase the threshold, generalization improves while locality declines gradually. Each model exhibits an optimal threshold where generalization and locality are balanced;  these thresholds can be adjusted to suit specific use cases e.g. high locality to ensure no degradation in the original model.

Figure \ref{fig:generalization_vs_locality} also illustrates the impact of varying the similarity threshold for edit-to-edit pairings in the training dataset on the projector network's learning. Edit-to-edit pairings $\phi$ which move edits farther away from each other are central to training a robust projector network. The threshold value for edit-to-edit pairings $\phi$ significantly impacts training stability and performance. Higher thresholds, such as 0.75, result in fewer pairings and lead to unstable training for both Llama-2-7b and GPT2-XL models, ultimately resulting in poor performance. Conversely, lower thresholds, exemplified by 0.6, increase the number of pairings and enhances stability.

\section{Downstream Task}
Table~\ref{tab:nlp_comparison} reports the performance of GPT2-XL and Llama-2-7B across downstream NLP tasks, before and after applying PENME. For GPT2-XL, performance remains unchanged post-edit, indicating that the edit was successfully localised without negatively affecting general capabilities. Similarly, Llama-2-7B exhibits stable performance on summarisation and classification tasks, with only a minor drop observed on the NLI task. These results suggest that PENME effectively preserves model general capabilities.

\section{Long Form Generation}
As discussed in Section~\ref{sec:penme_overview}, vector playback or trainable LoRA blocks can be used to support long-form generation. During the review process, an additional baseline was requested by the reviewers, which we subsequently included. In this section, we adopt the retrieval-based prompting strategy introduced by IKE~\citep{zheng2023can} . Since PENME operates on early layers of the model, inference can be halted early when user input falls within the scope of an edit. In such cases, the edited information is retrieved and used to construct a new prompt, combining the retrieved content with the user query to guide the model’s generation.

To evaluate this approach, metrics which include Generation Entropy (Fluency) and Consistency (Reference score)~\citep{meng2022mass} are used. Table~\ref{tab:generation_metrics} presents the results of this approach on the CounterFact dataset. We observe an improvement in fluency of the model's generation (fluency), indicating a reduction in repetitive or redundant output, and a corresponding increase in reference score, reflecting better factual alignment.

Sampled generations and detailed analysis is provided in Appendix~\ref{appendix:longform_samples}, the samples demonstrate that the edited fact is consistently preserved and integrated across the full output.

\section{Conclusion}
In this paper, we raised awareness of a critical vulnerability in weight-preserving adapter-based model editing techniques: lexical bias in the representation space. We developed a projection-based method PENME trained via contrastive learning to disentangle lexical and semantic similarity which originally would cause misfiring on irrelevant prompts with a high lexical overlap. 
Empirical evaluations showed PENME's superior performance across varying levels of task complexity. On the zsRE dataset, it achieved impressive generalisation and locality scores exceeding 0.90, demonstrating that our method is satisfactorily able to balance generalisation and locality using distance metrics in this new projected space. Notably, when assessed on the more challenging Counterfact benchmark, the system maintained robust performance, attaining scores above 0.80 for both generalisation and locality metrics. This performance on Counterfact is particularly significant given the benchmark's increased difficulty, underscoring PENME's efficacy. In future work, we aim to investigate whether a projector pretrained on a large-scale dataset can serve as a plug-and-play component for cross-lingual generalisation. Additionally, we plan to explore whether the projector can be trained and updated incrementally with new edits, thereby reducing training overhead and improving scalability.

\newpage
\section*{Impact Statement}
This paper advances model editing by mitigating lexical bias in adapter-based approaches, enabling precise and targeted updates to language models. While our method does not introduce an additional ethical risk beyond those already associated with language models, the model editing techniques, in general, can be exploited to inject unsafe behaviour into a model \citep{li2024badedit}.

\section*{Acknowledgment}
We acknowledge the support of the Natural Sciences and Engineering Research Council of Canada (NSERC), Canada Foundation of Innovation (CFI), and Research Nova Scotia. Advanced computing resources are provided by ACENET, the regional partner in Atlantic Canada, and the Digital Research Alliance of Canada.

\bibliography{example_paper}

\begin{thebibliography}{43}
\providecommand{\natexlab}[1]{#1}
\providecommand{\url}[1]{\texttt{#1}}
\expandafter\ifx\csname urlstyle\endcsname\relax
  \providecommand{\doi}[1]{doi: #1}\else
  \providecommand{\doi}{doi: \begingroup \urlstyle{rm}\Url}\fi

\bibitem[Dagan et~al.(2005)Dagan, Glickman, and Magnini]{dagan2005pascal}
Dagan, I., Glickman, O., and Magnini, B.
\newblock The pascal recognising textual entailment challenge.
\newblock In \emph{Machine learning challenges workshop}, pp.\  177--190. Springer, 2005.

\bibitem[Devlin et~al.(2019)Devlin, Chang, Lee, and Toutanova]{devlin2018bert}
Devlin, J., Chang, M., Lee, K., and Toutanova, K.
\newblock {BERT:} pre-training of deep bidirectional transformers for language understanding.
\newblock In Burstein, J., Doran, C., and Solorio, T. (eds.), \emph{Proceedings of the 2019 Conference of the North American Chapter of the Association for Computational Linguistics: Human Language Technologies, {NAACL-HLT} 2019, Minneapolis, MN, USA, June 2-7, 2019, Volume 1 (Long and Short Papers)}, pp.\  4171--4186. Association for Computational Linguistics, 2019.

\bibitem[Dumpala et~al.(2024)Dumpala, Jaiswal, Sastry, Milios, Oore, and Sajjad]{dumpala2024sugarcrepe++}
Dumpala, S.~H., Jaiswal, A., Sastry, C., Milios, E., Oore, S., and Sajjad, H.
\newblock {SUGARCREPE++ Dataset: Vision-Language Model Sensitivity to Semantic and Lexical Alterations}.
\newblock In \emph{{Conference on Neural Information Processing Systems, Dataset Track (NeurIPS)},}, 2024.

\bibitem[Fang et~al.(2025)Fang, Jiang, Wang, Ma, Shi, Wang, He, and Chua]{FangJWMSW0C25}
Fang, J., Jiang, H., Wang, K., Ma, Y., Shi, J., Wang, X., He, X., and Chua, T.
\newblock Alphaedit: Null-space constrained knowledge editing for language models.
\newblock In \emph{The Thirteenth International Conference on Learning Representations, {ICLR} 2025, Singapore, April 24-28, 2025}. OpenReview.net, 2025.

\bibitem[Geva et~al.(2021)Geva, Schuster, Berant, and Levy]{geva2020transformer}
Geva, M., Schuster, R., Berant, J., and Levy, O.
\newblock Transformer feed-forward layers are key-value memories.
\newblock In Moens, M., Huang, X., Specia, L., and Yih, S.~W. (eds.), \emph{Proceedings of the 2021 Conference on Empirical Methods in Natural Language Processing, {EMNLP} 2021, Virtual Event / Punta Cana, Dominican Republic, 7-11 November, 2021}, pp.\  5484--5495. Association for Computational Linguistics, 2021.

\bibitem[Gupta et~al.(2024{\natexlab{a}})Gupta, Rao, and Anumanchipalli]{gupta2024model}
Gupta, A., Rao, A., and Anumanchipalli, G.
\newblock Model editing at scale leads to gradual and catastrophic forgetting.
\newblock In Ku, L., Martins, A., and Srikumar, V. (eds.), \emph{Findings of the Association for Computational Linguistics, {ACL} 2024, Bangkok, Thailand and virtual meeting, August 11-16, 2024}, pp.\  15202--15232. Association for Computational Linguistics, 2024{\natexlab{a}}.

\bibitem[Gupta et~al.(2024{\natexlab{b}})Gupta, Sajnani, and Anumanchipalli]{gupta2024unified}
Gupta, A., Sajnani, D., and Anumanchipalli, G.
\newblock A unified framework for model editing.
\newblock In Al{-}Onaizan, Y., Bansal, M., and Chen, Y. (eds.), \emph{Findings of the Association for Computational Linguistics: {EMNLP} 2024, Miami, Florida, USA, November 12-16, 2024}, pp.\  15403--15418. Association for Computational Linguistics, 2024{\natexlab{b}}.

\bibitem[Hadsell et~al.(2006)Hadsell, Chopra, and LeCun]{hadsell2006dimensionality}
Hadsell, R., Chopra, S., and LeCun, Y.
\newblock Dimensionality reduction by learning an invariant mapping.
\newblock In \emph{2006 IEEE computer society conference on computer vision and pattern recognition (CVPR'06)}, volume~2, pp.\  1735--1742. IEEE, 2006.

\bibitem[Hartvigsen et~al.(2023)Hartvigsen, Sankaranarayanan, Palangi, Kim, and Ghassemi]{hartvigsen2024aging}
Hartvigsen, T., Sankaranarayanan, S., Palangi, H., Kim, Y., and Ghassemi, M.
\newblock Aging with {GRACE:} lifelong model editing with discrete key-value adaptors.
\newblock In Oh, A., Naumann, T., Globerson, A., Saenko, K., Hardt, M., and Levine, S. (eds.), \emph{Advances in Neural Information Processing Systems 36: Annual Conference on Neural Information Processing Systems 2023, NeurIPS 2023, New Orleans, LA, USA, December 10 - 16, 2023}, 2023.

\bibitem[Hase et~al.(2023)Hase, Bansal, Kim, and Ghandeharioun]{hase2023}
Hase, P., Bansal, M., Kim, B., and Ghandeharioun, A.
\newblock Does localization inform editing? surprising differences in causality-based localization vs. knowledge editing in language models.
\newblock In Oh, A., Naumann, T., Globerson, A., Saenko, K., Hardt, M., and Levine, S. (eds.), \emph{Advances in Neural Information Processing Systems}, volume~36, pp.\  17643--17668. Curran Associates, Inc., 2023.

\bibitem[Hermann et~al.(2015)Hermann, Kocisky, Grefenstette, Espeholt, Kay, Suleyman, and Blunsom]{hermann2015teaching}
Hermann, K.~M., Kocisky, T., Grefenstette, E., Espeholt, L., Kay, W., Suleyman, M., and Blunsom, P.
\newblock Teaching machines to read and comprehend.
\newblock \emph{Advances in neural information processing systems}, 28, 2015.

\bibitem[Huang et~al.(2024)Huang, Liu, Wang, and Liu]{huang2024reasons}
Huang, X., Liu, J., Wang, Y., and Liu, K.
\newblock Reasons and solutions for the decline in model performance after editing.
\newblock In Globersons, A., Mackey, L., Belgrave, D., Fan, A., Paquet, U., Tomczak, J.~M., and Zhang, C. (eds.), \emph{Advances in Neural Information Processing Systems 38: Annual Conference on Neural Information Processing Systems 2024, NeurIPS 2024, Vancouver, BC, Canada, December 10 - 15, 2024}, 2024.

\bibitem[Huang et~al.(2023)Huang, Shen, Zhang, Zhou, Rong, and Xiong]{huang2023transformer}
Huang, Z., Shen, Y., Zhang, X., Zhou, J., Rong, W., and Xiong, Z.
\newblock Transformer-patcher: One mistake worth one neuron.
\newblock In \emph{The Eleventh International Conference on Learning Representations, {ICLR} 2023, Kigali, Rwanda, May 1-5, 2023}. OpenReview.net, 2023.

\bibitem[Kwiatkowski et~al.(2019)Kwiatkowski, Palomaki, Redfield, Collins, Parikh, Alberti, Epstein, Polosukhin, Devlin, Lee, et~al.]{kwiatkowski2019natural}
Kwiatkowski, T., Palomaki, J., Redfield, O., Collins, M., Parikh, A., Alberti, C., Epstein, D., Polosukhin, I., Devlin, J., Lee, K., et~al.
\newblock Natural questions: a benchmark for question answering research.
\newblock \emph{Transactions of the Association for Computational Linguistics}, 7:\penalty0 453--466, 2019.

\bibitem[Levy et~al.(2017)Levy, Seo, Choi, and Zettlemoyer]{levy2017zero}
Levy, O., Seo, M., Choi, E., and Zettlemoyer, L.
\newblock Zero-shot relation extraction via reading comprehension.
\newblock In Levy, R. and Specia, L. (eds.), \emph{Proceedings of the 21st Conference on Computational Natural Language Learning (CoNLL 2017), Vancouver, Canada, August 3-4, 2017}, pp.\  333--342. Association for Computational Linguistics, 2017.

\bibitem[Li et~al.(2024{\natexlab{a}})Li, Deng, Cai, Lu, Chen, and Lam]{li2024comebahk}
Li, S., Deng, Y., Cai, D., Lu, H., Chen, L., and Lam, W.
\newblock Consecutive batch model editing with {H}oo{K} layers.
\newblock In Al-Onaizan, Y., Bansal, M., and Chen, Y.-N. (eds.), \emph{Proceedings of the 2024 Conference on Empirical Methods in Natural Language Processing}, pp.\  13817--13833, Miami, Florida, USA, November 2024{\natexlab{a}}. Association for Computational Linguistics.

\bibitem[Li et~al.(2024{\natexlab{b}})Li, Li, Song, Yang, Ma, and Yu]{li2024pmet}
Li, X., Li, S., Song, S., Yang, J., Ma, J., and Yu, J.
\newblock Pmet: Precise model editing in a transformer.
\newblock In \emph{Proceedings of the AAAI Conference on Artificial Intelligence}, pp.\  18564--18572, 2024{\natexlab{b}}.

\bibitem[Li et~al.(2024{\natexlab{c}})Li, Li, Chen, Zhang, Liu, Wang, Zhang, and Liu]{li2024badedit}
Li, Y., Li, T., Chen, K., Zhang, J., Liu, S., Wang, W., Zhang, T., and Liu, Y.
\newblock Badedit: Backdooring large language models by model editing.
\newblock In \emph{The Twelfth International Conference on Learning Representations, {ICLR} 2024, Vienna, Austria, May 7-11, 2024}. OpenReview.net, 2024{\natexlab{c}}.

\bibitem[Lin(2004)]{lin-2004-rouge}
Lin, C.-Y.
\newblock {ROUGE}: A package for automatic evaluation of summaries.
\newblock In \emph{Text Summarization Branches Out}, pp.\  74--81, Barcelona, Spain, July 2004. Association for Computational Linguistics.

\bibitem[Liu et~al.(2019)Liu, Ott, Goyal, Du, Joshi, Chen, Levy, Lewis, Zettlemoyer, and Stoyanov]{liu2019roberta}
Liu, Y., Ott, M., Goyal, N., Du, J., Joshi, M., Chen, D., Levy, O., Lewis, M., Zettlemoyer, L., and Stoyanov, V.
\newblock Roberta: A robustly optimized bert pretraining approach.
\newblock \emph{arXiv preprint arXiv:1907.11692}, 2019.

\bibitem[Luo et~al.(2023)Luo, Yang, Meng, Li, Zhou, and Zhang]{luo2023empirical}
Luo, Y., Yang, Z., Meng, F., Li, Y., Zhou, J., and Zhang, Y.
\newblock An empirical study of catastrophic forgetting in large language models during continual fine-tuning.
\newblock \emph{arXiv preprint arXiv:2308.08747}, 2023.

\bibitem[Ma et~al.(2025)Ma, Wang, Xu, Ling, and Gu]{MaWXLG25}
Ma, J., Wang, H., Xu, H., Ling, Z., and Gu, J.
\newblock Perturbation-restrained sequential model editing.
\newblock In \emph{The Thirteenth International Conference on Learning Representations, {ICLR} 2025, Singapore, April 24-28, 2025}. OpenReview.net, 2025.

\bibitem[Meng et~al.(2022)Meng, Bau, Andonian, and Belinkov]{meng2022locating}
Meng, K., Bau, D., Andonian, A., and Belinkov, Y.
\newblock Locating and editing factual associations in {GPT}.
\newblock In Koyejo, S., Mohamed, S., Agarwal, A., Belgrave, D., Cho, K., and Oh, A. (eds.), \emph{Advances in Neural Information Processing Systems 35: Annual Conference on Neural Information Processing Systems 2022, NeurIPS 2022, New Orleans, LA, USA, November 28 - December 9, 2022}, 2022.

\bibitem[Meng et~al.(2023)Meng, Sharma, Andonian, Belinkov, and Bau]{meng2022mass}
Meng, K., Sharma, A.~S., Andonian, A.~J., Belinkov, Y., and Bau, D.
\newblock Mass-editing memory in a transformer.
\newblock In \emph{The Eleventh International Conference on Learning Representations, {ICLR} 2023, Kigali, Rwanda, May 1-5, 2023}. OpenReview.net, 2023.

\bibitem[Miller et~al.(2016)Miller, Fisch, Dodge, Karimi, Bordes, and Weston]{miller2016key}
Miller, A.~H., Fisch, A., Dodge, J., Karimi, A., Bordes, A., and Weston, J.
\newblock Key-value memory networks for directly reading documents.
\newblock In Su, J., Carreras, X., and Duh, K. (eds.), \emph{Proceedings of the 2016 Conference on Empirical Methods in Natural Language Processing, {EMNLP} 2016, Austin, Texas, USA, November 1-4, 2016}, pp.\  1400--1409. The Association for Computational Linguistics, 2016.

\bibitem[Mitchell et~al.(2021)Mitchell, Lin, Bosselut, Finn, and Manning]{mitchell2021fast}
Mitchell, E., Lin, C., Bosselut, A., Finn, C., and Manning, C.~D.
\newblock Fast model editing at scale.
\newblock In \emph{International Conference on Learning Representations}, 2021.

\bibitem[Mitchell et~al.(2022)Mitchell, Lin, Bosselut, Manning, and Finn]{mitchell2022memory}
Mitchell, E., Lin, C., Bosselut, A., Manning, C.~D., and Finn, C.
\newblock Memory-based model editing at scale.
\newblock In \emph{International Conference on Machine Learning}, pp.\  15817--15831. PMLR, 2022.

\bibitem[Radford et~al.(2019)Radford, Wu, Child, Luan, Amodei, Sutskever, et~al.]{radford2019language}
Radford, A., Wu, J., Child, R., Luan, D., Amodei, D., Sutskever, I., et~al.
\newblock Language models are unsupervised multitask learners.
\newblock \emph{OpenAI blog}, 1\penalty0 (8):\penalty0 9, 2019.

\bibitem[Raffel et~al.(2020)Raffel, Shazeer, Roberts, Lee, Narang, Matena, Zhou, Li, and Liu]{raffel2020exploring}
Raffel, C., Shazeer, N., Roberts, A., Lee, K., Narang, S., Matena, M., Zhou, Y., Li, W., and Liu, P.~J.
\newblock Exploring the limits of transfer learning with a unified text-to-text transformer.
\newblock \emph{Journal of machine learning research}, 21\penalty0 (140):\penalty0 1--67, 2020.

\bibitem[Rosati et~al.(2024)Rosati, Gonzales, Chen, Yu, Kayani, Rudzicz, and Sajjad]{rosati-etal-2024-long}
Rosati, D., Gonzales, R., Chen, J., Yu, X., Kayani, Y., Rudzicz, F., and Sajjad, H.
\newblock Long-form evaluation of model editing.
\newblock In Duh, K., Gomez, H., and Bethard, S. (eds.), \emph{Proceedings of the 2024 Conference of the North American Chapter of the Association for Computational Linguistics: Human Language Technologies (Volume 1: Long Papers)}, pp.\  3749--3780, Mexico City, Mexico, June 2024. Association for Computational Linguistics.

\bibitem[Saravia et~al.(2018)Saravia, Liu, Huang, Wu, and Chen]{saravia-etal-2018-carer}
Saravia, E., Liu, H.-C.~T., Huang, Y.-H., Wu, J., and Chen, Y.-S.
\newblock {CARER}: Contextualized affect representations for emotion recognition.
\newblock In \emph{Proceedings of the 2018 Conference on Empirical Methods in Natural Language Processing}, pp.\  3687--3697, Brussels, Belgium, October-November 2018. Association for Computational Linguistics.

\bibitem[Takase et~al.(2022)Takase, Kiyono, Kobayashi, and Suzuki]{takase2022layer}
Takase, S., Kiyono, S., Kobayashi, S., and Suzuki, J.
\newblock On layer normalizations and residual connections in transformers.
\newblock \emph{arXiv preprint arXiv:2206.00330}, 2022.

\bibitem[Touvron et~al.(2023{\natexlab{a}})Touvron, Lavril, Izacard, Martinet, Lachaux, Lacroix, Rozière, Goyal, Hambro, Azhar, Rodriguez, Joulin, Grave, and Lample]{Llama}
Touvron, H., Lavril, T., Izacard, G., Martinet, X., Lachaux, M.-A., Lacroix, T., Rozière, B., Goyal, N., Hambro, E., Azhar, F., Rodriguez, A., Joulin, A., Grave, E., and Lample, G.
\newblock Llama: Open and efficient foundation language models, 2023{\natexlab{a}}.

\bibitem[Touvron et~al.(2023{\natexlab{b}})Touvron, Martin, Stone, Albert, Almahairi, Babaei, Bashlykov, Batra, Bhargava, Bhosale, et~al.]{touvron2023Llama}
Touvron, H., Martin, L., Stone, K., Albert, P., Almahairi, A., Babaei, Y., Bashlykov, N., Batra, S., Bhargava, P., Bhosale, S., et~al.
\newblock Llama 2: Open foundation and fine-tuned chat models.
\newblock \emph{arXiv preprint arXiv:2307.09288}, 2023{\natexlab{b}}.

\bibitem[Wang et~al.(2023)Wang, Zhang, Xie, Yao, Tian, Wang, Xi, Cheng, Liu, Zheng, et~al.]{wang2023easyedit}
Wang, P., Zhang, N., Xie, X., Yao, Y., Tian, B., Wang, M., Xi, Z., Cheng, S., Liu, K., Zheng, G., et~al.
\newblock Easyedit: An easy-to-use knowledge editing framework for large language models.
\newblock \emph{arXiv preprint arXiv:2308.07269}, 2023.

\bibitem[Wang et~al.(2024)Wang, Li, Zhang, Xu, Yao, Jiang, Xie, Huang, and Chen]{wang2024wise}
Wang, P., Li, Z., Zhang, N., Xu, Z., Yao, Y., Jiang, Y., Xie, P., Huang, F., and Chen, H.
\newblock {WISE:} rethinking the knowledge memory for lifelong model editing of large language models.
\newblock In Globersons, A., Mackey, L., Belgrave, D., Fan, A., Paquet, U., Tomczak, J.~M., and Zhang, C. (eds.), \emph{Advances in Neural Information Processing Systems 38: Annual Conference on Neural Information Processing Systems 2024, NeurIPS 2024, Vancouver, BC, Canada, December 10 - 15, 2024}, 2024.

\bibitem[Wang et~al.(2025)Wang, Zhu, Liu, Zheng, Chen, and Li]{wang2023knowledge}
Wang, S., Zhu, Y., Liu, H., Zheng, Z., Chen, C., and Li, J.
\newblock Knowledge editing for large language models: {A} survey.
\newblock \emph{{ACM} Comput. Surv.}, 57\penalty0 (3):\penalty0 59:1--59:37, 2025.

\bibitem[Yao et~al.(2023)Yao, Wang, Tian, Cheng, Li, Deng, Chen, and Zhang]{yao2023editing}
Yao, Y., Wang, P., Tian, B., Cheng, S., Li, Z., Deng, S., Chen, H., and Zhang, N.
\newblock Editing large language models: Problems, methods, and opportunities.
\newblock In Bouamor, H., Pino, J., and Bali, K. (eds.), \emph{Proceedings of the 2023 Conference on Empirical Methods in Natural Language Processing, {EMNLP} 2023, Singapore, December 6-10, 2023}, pp.\  10222--10240. Association for Computational Linguistics, 2023.

\bibitem[Yu et~al.(2024)Yu, Chen, Zhou, and He]{yu2024melo}
Yu, L., Chen, Q., Zhou, J., and He, L.
\newblock Melo: Enhancing model editing with neuron-indexed dynamic lora.
\newblock In \emph{Proceedings of the AAAI Conference on Artificial Intelligence}, pp.\  19449--19457, 2024.

\bibitem[Zheng et~al.(2023)Zheng, Li, Dong, Fan, Wu, Xu, and Chang]{zheng2023can}
Zheng, C., Li, L., Dong, Q., Fan, Y., Wu, Z., Xu, J., and Chang, B.
\newblock Can we edit factual knowledge by in-context learning?
\newblock In Bouamor, H., Pino, J., and Bali, K. (eds.), \emph{Proceedings of the 2023 Conference on Empirical Methods in Natural Language Processing, {EMNLP} 2023, Singapore, December 6-10, 2023}, pp.\  4862--4876. Association for Computational Linguistics, 2023.

\bibitem[Zhou \& Bansal(2020)Zhou and Bansal]{zhou-bansal-2020-towards}
Zhou, X. and Bansal, M.
\newblock Towards robustifying {NLI} models against lexical dataset biases.
\newblock In Jurafsky, D., Chai, J., Schluter, N., and Tetreault, J. (eds.), \emph{Proceedings of the 58th Annual Meeting of the Association for Computational Linguistics}, pp.\  8759--8771, Online, July 2020. Association for Computational Linguistics.

\bibitem[Zhu et~al.(2020)Zhu, Rawat, Zaheer, Bhojanapalli, Li, Yu, and Kumar]{zhu2020modifying}
Zhu, C., Rawat, A.~S., Zaheer, M., Bhojanapalli, S., Li, D., Yu, F., and Kumar, S.
\newblock Modifying memories in transformer models.
\newblock \emph{arXiv preprint arXiv:2012.00363}, 2020.

\bibitem[Zhu et~al.(2024)Zhu, Lan, Li, and Qian]{zhu2024initializing}
Zhu, H., Lan, Y., Li, X., and Qian, W.
\newblock Initializing and retrofitting key-value adaptors for traceable model editing, 2024.
\newblock OpenReview preprint.

\end{thebibliography}
\bibliographystyle{icml2025}

%%%%%%%%%%%%%%%%%%%%%%%%%%%%%%%%%%%%%%%%%%%%%%%%%%%%%%%%%%%%%%%%%%%%%%%%%%%%%%%
%%%%%%%%%%%%%%%%%%%%%%%%%%%%%%%%%%%%%%%%%%%%%%%%%%%%%%%%%%%%%%%%%%%%%%%%%%%%%%%
% APPENDIX
%%%%%%%%%%%%%%%%%%%%%%%%%%%%%%%%%%%%%%%%%%%%%%%%%%%%%%%%%%%%%%%%%%%%%%%%%%%%%%%
%%%%%%%%%%%%%%%%%%%%%%%%%%%%%%%%%%%%%%%%%%%%%%%%%%%%%%%%%%%%%%%%%%%%%%%%%%%%%%%
\newpage
\appendix
\onecolumn

% You can have as much text here as you want. The main body must be at most $8$ pages long.
% For the final version, one more page can be added.
% If you want, you can use an appendix like this one.  

% The $\mathtt{\backslash onecolumn}$ command above can be kept in place if you prefer a one-column appendix, or can be removed if you prefer a two-column appendix.  Apart from this possible change, the style (font size, spacing, margins, page numbering, etc.) should be kept the same as the main body.

\section{Data Construction and Inference for PENME}
\label{sec:penme-details}
The projection network is similar to the feed-forward layers in a transformer as it contains
two layers with ReLU activation in between, with the addition of the Batch Normalization layer, a common
element in contrastive learning. The network is trained via contrastive learning, which requires a dataset based on a pair of inputs with positive and negative labels. The algorithm \ref{algo:data_construction} data construction process. 

At runtime, upon receiving a user query, PEMNE checks if the query falls within the editing scope of the edits. If so the new output from memory is retrieved and the inference process is stopped. Alternative generation mechanism is discussed in Section~\ref{sec:penme_overview} of the main paper text, which includes replacing the storage of the new fact as text with playback vectors or LoRa block indices. Additionally, using the retrieved information in context learning based generation can be used. The inference pipeline for PENME is given in \ref{algo:penme}.

\begin{algorithm}[ht]
  \caption{Data Construction for Projector Network}
  \label{algo:data_construction}
  \begin{algorithmic}[1]
    \STATE {\bfseries Input:} $num\_overall\_negative$, $threshold\_edit\_pairings$
    \STATE {\bfseries Input:} memory $\gets \{\}$ \hfill\COMMENT{Memory storage}
    \STATE {\bfseries Input:} dataset\_pairs $\gets []$
    \STATE {\bfseries Input:} $Cos(\cdot, \cdot)$ \hfill\COMMENT{Cosine similarity function}
    \STATE {\bfseries Input:} dataset rows $r_i = (x_i, y_i, \{p_{ij}\}, \{p^{\neg}_{ij}\})$

    \FOR{each $r_i$ in dataset}
      \FOR{each $p_{ij}$ and $p^{\neg}_{ij}$ in $r_i$}
        \STATE Add $(x_i, p_{ij})$ to dataset\_pairs \hfill\COMMENT{Positive pair}
        \STATE Add $(x_i, p^{\neg}_{ij})$ to dataset\_pairs \hfill\COMMENT{Negative pair}
      \ENDFOR

      \FOR{each $r_t$ in dataset, where $i \neq t$}
        \IF{$Cos(x_i, x_t) > threshold\_edit\_pairings$}
          \STATE Add $(x_i, x_t)$ to dataset\_pairs \hfill\COMMENT{Negative edit-to-edit pair}
        \ENDIF

        \FOR{each $p^{\neg}_{tj}$ in $r_t$}
          \STATE Add $(Cos(x_i, p^{\neg}_{tj}), (x_i, p^{\neg}_{tj}))$ to memory
        \ENDFOR
      \ENDFOR
    \ENDFOR

    \STATE Sort memory in descending order by similarity
    \STATE Add top-$num\_overall\_negative$ items from memory to dataset\_pairs
    \STATE \textbf{return} dataset\_pairs
  \end{algorithmic}
\end{algorithm}

% \begin{algorithm*}
%   \caption{Inference for LLM with PENME}
%   \begin{algorithmic}[1]
%   \State \textbf{Input} $h_l(\cdot)$$ \Comment{LLM model output at layer l}
%   \State \textbf{Input} $g(\cdot)$ \Comment{Projector network}
%   \State \textbf{Input} $D(\cdot,\cdot)$ \Comment{Euclidean Distance function}
%   \State \textbf{Input} codebook = \{(keys(K)=vectors, values(V)=(threshold, output))\}
%     \State \textbf{Input} $x_t$ user prompt2
%     % \State \textbf{Input} dataset rows $x_t = (edit_t,[paraphrase_{t0}...paraphrase_{tn}],[neighbour_{t0}...neighbour_{tn}])$\;
%     \State $h_l$ $\leftarrow  h_l(x_t)$ \;
%     \State $act_x$ $\leftarrow  g(h_l)$ \;
    
%     \State $\text{selectedKey} \leftarrow$ $min_i(D(act_x,k_i))$
%     \Comment{compute Euclidean distance between $act_x$ and keys in codebook and extract closet key}
%      \If{$D(act_x, selectedKey) < V[selectedKey][threshold]$}\;
%         \State \Return $V[selectedKey][output]$
%      \EndIf
%     \State \Return y
%   \end{algorithmic}
%   \label{algo:penme}
% \end{algorithm*}

\begin{algorithm}[ht]
  \caption{Inference for LLM with PENME}
  \label{algo:penme}
  \begin{algorithmic}[1]
    \STATE \textbf{Input:} $h_l(\cdot)$ \hfill\COMMENT{LLM output at layer $l$}
    \STATE \textbf{Input:} $g(\cdot)$ \hfill\COMMENT{Projector network}
    \STATE \textbf{Input:} $D(\cdot, \cdot)$ \hfill\COMMENT{Euclidean distance}
    \STATE \textbf{Input:} codebook = \{$k_i$: ($v_i^{\delta}$ , $v_i^{\mu}$ )\}
    \STATE \textbf{Input:} user prompt $x_t$

    \STATE $h_l \leftarrow h_l(x_t)$
    \STATE $a_x \leftarrow g(h_l)$

    \STATE Find $k^* = \arg\min_{k_i} D(a_x, k_i)$
    \IF{$D(a_x, k^*) < v_{k^*}^{\delta} $}
        \STATE \textbf{return} $y_{k^*}$
    \ELSE
        \STATE \textbf{return} $y$ \hfill\COMMENT{Fallback to base LLM output}
    \ENDIF
  \end{algorithmic}
\end{algorithm}

% Avg Paraphrase Distance (AvgPD)
% Min Paraphrase Distance (MaxPD) (MinPD)
% Avg Distance Closest Paraphrase and Farthest Neighbour (AvgCPFN
\section{Paraphrases and Irrelevant prompts Distance Analysis}
\label{sec:pre_post_proj}
Table~\ref{tab:edit_paraphrase_neighbour_dist} shows the distance between edits and their respective paraphrases and irrelevant prompts across various measurement metrics. From the distances the average paraphrase distance (AvgPD) and average distances between training and test paraphrases (AvgDTTP), we can see that they are generally a little farther than the test paraphrases and are on average a bit farther from the edit than train paraphrases. On the other hand, the average irrelevant prompt distance (AvgPN) and average distances between training and test irrelevant prompts (AvgDTTN) show that the test irrelevant prompts are a little closer to the edit as compared to the train irrelevant prompts.

\begin{table*}[ht]
\caption{Distance analysis of distances between edit and its respective paraphrase and irrelevant prompts.  The metrics for measurement include average/max/min paraphrase distance (AvgPD)(MaxPD)(MinPD), average/max/min irrelevant prompts distance (AvgND),(MaxND)(MinND), average/max/min distances between training and test paraphrase (AvgDTTP)(MaxDTTP)(MinDTTP), the average distance between farthest edit and closest irrelevant prompt (AvgCPFN) and average/max/min distances between training and test irrelevant prompts (AvgDTTN)(MaxDTTN)(MinDTTN)}
    \centering
    \scriptsize
    \renewcommand{\arraystretch}{1.2}
    \begin{tabularx}{\textwidth}{X X X X }
    % \textwidth}{p{0.14\textwidth} p{0.14\textwidth} >{\centering\arraybackslash}p{0.14\textwidth} >{\centering\arraybackslash}p{0.14\textwidth} >{\centering\arraybackslash}p{0.14\textwidth} >{\centering\arraybackslash}p{0.14\textwidth}}
    
        \hline
        \textbf{Model} & \textbf{Measurement Metric} & \textbf{ Training Set} & \textbf{Test Set}   \\
        \hline

         \multirow{13}{*}{\textbf{Llama-2-7b}} & AvgPD & 0.240 & 0.254\\
         & MinPD &  0.0  & 0.02 \\
         & MaxPD &   0.829 &   1.59\\
         & AvgND &   1.436  &   1.379\\
         & MinND &  0.803  &  0.616 \\
         & MaxND &  1.884  &  1.853 \\
         & AvgCPFN &  0.348  &  0.893\\
         \cline{3-4}
            &  & \multicolumn{2}{c}{\textbf{Training Set vs Test Set }}  \\
        \cline{3-4}
         & AvgDTTP & \multicolumn{2}{c}{0.013} \\
         & MaxDTTP & \multicolumn{2}{c}{1.459} \\
         & MinDTTP & \multicolumn{2}{c}{-0.634} \\
         & AvgDTTN & \multicolumn{2}{c}{-0.227} \\
         & MaxDTTN & \multicolumn{2}{c}{-1.130} \\
         & MinDTTN & \multicolumn{2}{c}{0.0} \\
        \hline
         \multirow{7}{*}{\textbf{T5-small}} & AvgPD & 0.409 & 0.491\\
         & MinPD &  0.0  &  0.002\\
         & MaxPD &   1.375 &   1.381\\
         & AvgND &   0.468  &   0.534\\
         & MinND &  0.005 &  0.010 \\
         & MaxND &   1.384 &  1.386 \\
         & AvgCPFN & 0.193 &  0.238\\
          \cline{3-4}
            &  & \multicolumn{2}{c}{\textbf{Training Set vs Test Set }}  \\
        \cline{3-4}
         & AvgDTTP & \multicolumn{2}{c}{0.018} \\
         & MaxDTTP & \multicolumn{2}{c}{1.273} \\
         & MinDTTP & \multicolumn{2}{c}{-1.290} \\
         & AvgDTTN & \multicolumn{2}{c}{-0.276 } \\
         & MaxDTTN & \multicolumn{2}{c}{-1.341} \\
         & MinDTTN & \multicolumn{2}{c}{0.0} \\
        \hline
        \multirow{7}{*}{\textbf{GPT2-XL}} & AvgPD & 0.378 & 0.349\\
        & MinPD &  0.0  &   0.01\\
         & MaxPD &   1.49 &   1.395\\
         & AvgND &   1.174  &   1.092\\
         & MinND &  0.227  &  0.368 \\
         & MaxND &  1.709  &  1.728 \\
         & AvgCPFN &  0.382  &  0.700\\
          \cline{3-4}
            &  & \multicolumn{2}{c}{\textbf{Training Set vs Test Set }}  \\
        \cline{3-4}
         & AvgDTTP & \multicolumn{2}{c}{0.008} \\
         & MaxDTTP & \multicolumn{2}{c}{1.368} \\
         & MinDTTP & \multicolumn{2}{c}{-1.046} \\
         & AvgDTTN & \multicolumn{2}{c}{-0.148 } \\
         & MaxDTTN & \multicolumn{2}{c}{-0.856} \\
         & MinDTTN & \multicolumn{2}{c}{0.0} \\
        \hline
        \hline
    \end{tabularx}
    
    \label{tab:edit_paraphrase_neighbour_dist}
    \vspace{-0.4cm}
\end{table*}

\section{Comparison Scoping Mechanism: PENME versus MELO and GRACE}
\label{sec:penme_vs_grace_melo_scope}
To demonstrate the improvement in inference time for selecting the appropriate key, we compare PENME with MELO across various sample sizes of edits, ranging from 50 to 300 in increments of 50 shown in table \ref{tab:runtime_comparison}. The results show that PENME outperforms MELO in terms of speed and also highlight the number of keys forgotten during training due to the design of its scoping mechanism, as well as the number of entries for which the radius had to be reduced.

\begin{table*}
\caption{Runtime Performance Comparison of PENME versus MELO. For PENME the number of Codebook entries is the same as the number of edits.}
    \centering
    \scriptsize
    
    \begin{tabularx}{\textwidth}{X c | c c c c}
        \toprule
        \textbf{Number of Edits} & {\textbf{PENME}} & \multicolumn{4}{c}{\textbf{MELO/GRACE}}  \\
        \midrule 
        & \textbf{Runtime (ms)} & \textbf{Runtime (ms)}& \textbf{Codebook Entries} & \textbf{Edits Forgotten} & \textbf{Edit Conflict} \\
        \midrule
        50  &  0.024 ± 0.003  & 0.316 ± 0.090 & 269 & 24 & 21  \\
        100  & 0.115 ± 0.129 & 0.364 ± 0.050 & 523 & 77  & 66    \\
        150 & 0.188 ± 0.182 & 0.624 ± 0.082& 785 & 132 & 114   \\
        200 & 0.279 ± 0.170  & 1.423 ± 0.180 & 1048 & 188 & 169  \\
        250 & 0.404 ± 0.170 & 1.681 ± 0.205& 1319  & 254 & 217   \\
        300 & 0.418 ± 0.125 & 2.149 ± 1.069 & 1554  & 301  & 268 \\
        \bottomrule
    \end{tabularx}
    
    \label{tab:runtime_comparison}
\end{table*}

\section{Experimentation and Implementation Details}

\subsection{Experimentation Setup}
\label{sec:experimental-setup}
For our comparative analysis, we contrast against baseline methods such as simple fine-tuning (FT), alongside advanced approaches drawn from relevant literature. These encompass GRACE \citep{hartvigsen2024aging,yu2024melo}, employing adapter-based editing with a similarity-based scoping mechanism. SERAC \citep{mitchell2022memory}, a multimodal editing approach incorporating a scoping classifier, memory database, and counterfactual model alongside the target model and MEMIT \citep{meng2022mass} an editing approach designed for decoder only model adopts a model-editing strategy by identifying and updating knowledge-contained model layers' weight matrices.

In evaluating our approach, we adhere to the metrics outlined in \S \ref{sec:prob_dataset}. Regarding generalization, we define a paraphrase as generalized if it aligns with the correct edit and falls below its distance threshold. For assessing locality, we maintain that locality is preserved when the distance between matched edits exceeds its threshold. Any other instances are categorized as misfires. It is important to note that \citep{hartvigsen2024aging,yu2024melo} utilize token F1 Accuracy and \citep{mitchell2022memory} use a metric based on token probabilities. These metrics are softer in nature which allows for higher scores.

\subsubsection{Computation Resources}
Training for all projector networks is conducted on an NVIDIA P100 GPU with 16GB VRAM. A larger VRAM or RAM capacity is only necessary for the initial extraction of layer representations from the pre-trained language models. For the evaluation of approaches from relevant literature, some of which demanded greater computational resources, we employed  NVIDIA A100 GPUs with 40GB and 80GB VRAM. All editing approaches were supported are implemented using the default configurations provided in the Easy-Editor library \citep{wang2023easyedit}. It is important to note that not all models are supported across all editing methods. For instance, Llama-2-7b is not supported for MELO. For some models such as T5-small, limited support is provided therefore, we utilise the code provided by the paper's authors.

\subsubsection{Hyperparameters}
For training projector networks we utilize the Adam optimizer. we experiment with various learning rates $1e^{1-2}$, $2e^{1-2}$, $3e^{1-2}$. we find that a moderate learning rate is required to learn faster while not overfitting, hence we choose $1e^{1-2}$, with a learning rate decay rate of $0.01$. All projection networks are trained for 200 epochs using a batch size of $8192$ and an early stopping patience of 8 epochs. For selecting the margin $m$ in the contrastive learning cost function we ablate on the hyperparameter m for the GPT2-XL model. The table \ref{tab:abalation_m} shows the margin m along with the adjustment to $\tau$ for balanced results for generalization and locality. It can be observed from the table to achieve high-performance minimum value of $30$ needs to be utilized. The higher the the value for $m$ the better the score for localization. The value chosen is 40 which has the most balanced results.
\begin{table*}[ht]
\caption{The table shows how the performance changes along with the required threshold adjustment to $\tau$ as margin $m$ in contrastive loss is changed}
    \centering
    \scriptsize
    
    \begin{tabularx}{\textwidth}{X X X X}
        \toprule
        \textbf{{Margin \textbf{$m$}}} & \textbf{Threshold Adjustment} {\textbf{$\tau$}} & \textbf{Generalization} & \textbf{Locality}\\
 
        \midrule
        10 & 0 & 0.634  & 0.831    \\
        20 & 3 & 0.891 & 0.880 \\
        30 & 6 & 0.958 & 0.948\\
        40 & 8 & 0.967 & 0.977 \\
        50 & 11 & \textbf{0.978 }& 0.965 \\
        60 & 13 & 0.976 & 0.986\\
        70 & 17 & 0.973 & 0.976\\
        80 & 17 & 0.973 & 0.976\\
        90 & 20 & 0.928 & \textbf{0.986}\\

        \bottomrule
    \end{tabularx}
    
    \label{tab:abalation_m}
\end{table*}
% Due to computational resource constraints, the experimental design could not be fully executed. Specifically, experiments were limited to 200 samples of the Counterfact dataset. Furthermore, the absence of large GPUs necessitated the omission of experiments involving GRACE on Llama-2-7b and GPT2-XL, as well as MEMIT on Llama-2-7b. Additionally, MEMIT experiments on the T5 model were excluded because the approach is compatible only with decoder-based architectures. 

\subsection{Data Processing }
\label{sec:counterfact-data-processing}
\textbf{Counterfact:} Each row in the Counterfact consists of an edit prompt, two paraphrase prompts, multiple irrelevant prompts and an edit label $x_i,y_i,[p_{1},p_{2}],[p^{\neg}_{i1}...p^{\neg}_{ij}])$. For the training dataset, we extract the edit prompt $x_i$, one randomly sampled paraphrase $p_i$ and half the irrelevant prompts $p^{\neg}_{ij}$. For creating additional paraphrases for the training set we utilize the extracted edit prompt and paraphrase prompt as input to ChatGPT and use it to generate three additional paraphrases for training. We ensure that the generated paraphrase follows the $(s,r,o^*)$ triplet format that the dataset uses. The test set for locality and generalization compromises of the paraphrase and irrelevant prompts not sampled from the training set.

\textbf{zsRE:} The zsRE dataset comprises of rows containing a sample question, its corresponding new label, and multiple rephrased questions along with its filtered rephrased questions. We constructed this dataset following methodologies established in the relevant literature. A balanced subset of paraphrases are derived from the filtered rephrased questions for training and testing purposes. For irrelevant prompts samples, we randomly selected an equal number of questions from the NQ dataset for training and testing while ensuring no overlap in questions.

To highlight the lexicality issue in the datasets, we compute several token overlap metrics between pairs of (edits, paraphrases) $(x_i, p_{ij})$ and (edits, irrelevant prompts) $(x_i, p^{\neg}_{ij})$. The results are presented in Table~\ref{tab:token_metrics} and dataset samples in Table~\ref{tab:data_samples}. From the token overlap metrics table, it is evident that the edit prompt and irrelevant prompts show high overlap in Counterfact, whereas the overlap is minimal in ZsRE. This, coupled with the experiment in Section~\ref{sec:lexical-dominance}, highlights the significant challenges observed in the Counterfact dataset.

\begin{table*}[ht]
\caption{Comparison between ZsRE and Counterfact for token overlap metrics}
    \centering
    \scriptsize
    
    \begin{tabularx}{\textwidth}{c c c c c c | c c c  c}
        \toprule
         & \multicolumn{4}{c}{\textbf{ZsRE}} & \multicolumn{4}{c}{\textbf{Counterfact}}  \\
        
        \midrule 
        
        \textbf{Metric} & \textbf{Pair Type} & \textbf{Score} & \textbf{Precision} & \textbf{Recall} & \textbf{F1}  &  \textbf{Value} & \textbf{Precision} & \textbf{Recall} & \textbf{F1}\\
        \midrule
        Jaccard Similarity  &  $(x_i,p_{ij})$  & 0.399  & - & - & - & 0.401 & - & - & - \\
        Jaccard Similarity  &  $(x_i,p^{\neg}_{ij})$  & 0.086 & - & - & - & 0.430 & - & - & - \\
        ROUGE-1  &  $(x_i,p_{ij})$  & - & 0.321 & 0.315  & 0.316 & - & 0.310 & 0.325 & 0.307 \\
        ROUGE-1  &  $(x_i,p^{\neg}_{ij})$ & - &  0.076 & 0.087 & 0.079 & - & 0.295 & 0.293 & 0.290 \\
        ROUGE-2  &  $(x_i,p_{ij})$  & -  & 0.189 & 0.194 & 0.194 & - & 0.189 & 0.198 & 0.184 \\
        ROUGE-2  &  $(x_i,p^{\neg}_{ij})$  & - & 0.008 & 0.008 & 0.008 & - & 0.205 & 0.203 & 0.201 \\
        ROUGE-L  &  $(x_i,p_{ij})$  & -  & 0.299 & 0.294 & 0.293 &-& 0.299 & 0.312 & 0.295 \\
        ROUGE-L &  $(x_i,p^{\neg}_{ij})$  & - & 0.070 & 0.080 &  0.073 & - & 0.294 & 0.292 & 0.289 \\
        
        \bottomrule
    \end{tabularx}
    
    \label{tab:token_metrics}
\end{table*}

\begin{table*}[ht]
    \centering
    \footnotesize
    
    \begin{tabularx}{\textwidth}{X X X | X X X}
        \toprule
         \multicolumn{3}{c}{\textbf{Counterfact}} & \multicolumn{3}{c}{\textbf{ZsRE}}  \\
        \midrule 
        \textbf{Edit} & \textbf{Paraphrase} & \textbf{Neighbour} & \textbf{Edit} & \textbf{Paraphrase} & \textbf{Neighbour NQ dataset} \\
        \midrule
           The twin city of Cologne is & What is the twin city of Cologne? It is  & The twin city of London is & Which river system contains Laborec? & What river system does Laborec contain? & Where does the last name serrano come from?  \\
           Alexander Zinoviev works in the area of & Alexander Zinoviev's domain of work is  & TFred W. Riggs works in the area of & Which airport does Air Seychelles operate in? & Which airport is closely linked to Air Seychelles? & How many students attend chippewa valley high school?  \\
           The original language of Kondura was & The language of Kondura is  & The language of Taal is & The country of origin for Kala Pul is what? & Which was the country for Kala Pul? & "When do the new sky sports channels launch?  \\
           Thomas Arne died in the city of & Thomas Arne lost their life at  & Bill Brandt died in the city of & What label was responsible for Wild World? & What was the label Wild World? & Who composed the music for avengers infinity war?  \\
        \bottomrule
    \end{tabularx}
    \caption{Random samples from the Counterfact and ZsRE datasets.}
    \label{tab:data_samples}
\end{table*}

\vfill
\pagebreak

\section{Projector Network and Lexical Dominance}

\subsection{Lexical Dominance Layer Analysis}
\label{sec:lexical_dominance_details}
\begin{figure*}[ht]
    % \centering
    \includegraphics[width=1\textwidth]{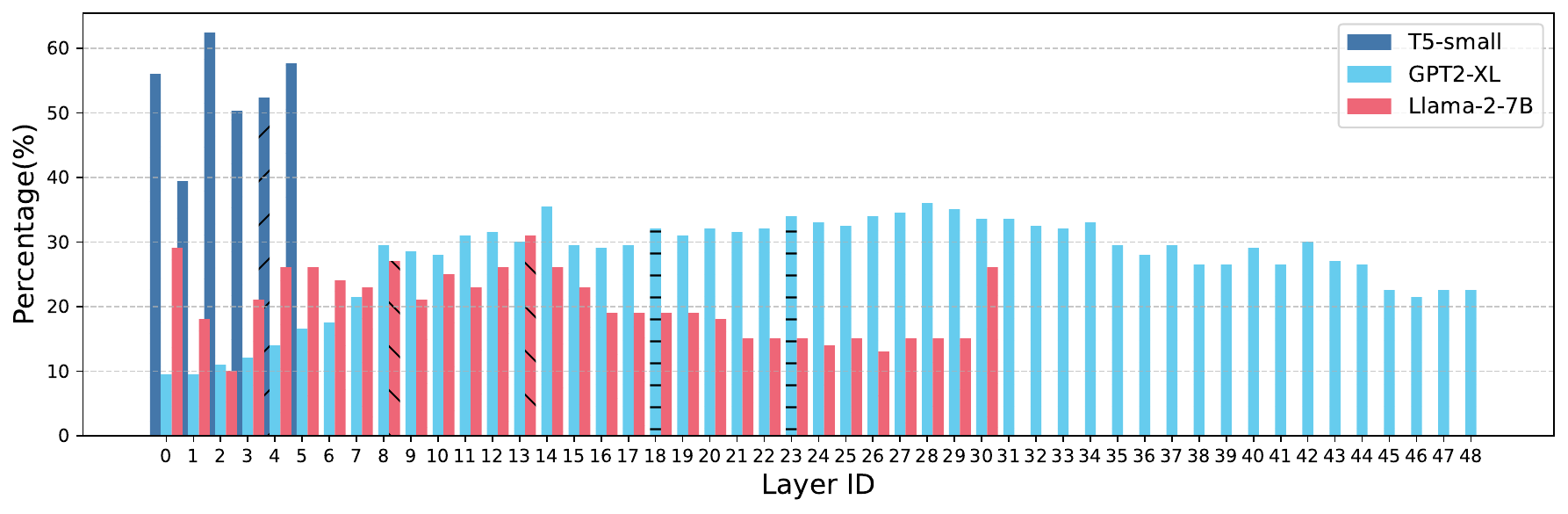} % Change 
    \caption{Percentage of samples \textit{where edits are closer} to irrelevant prompts as compared to paraphrases in the representations space of different models across all layers.  T5-small, GPT2-XL and Llama-2-7b have 6, 32, 48 layers, respectively.}
    \label{fig:lexical_dominance_complete}

\end{figure*}

Figure \ref{fig:lexical_dominance_complete} shows the percentage of edits
samples where irrelevant prompts were closer to the edits for all models across all layers.

\subsection{Layer-Wise Analysis of the Projector Network}
\label{sec:layer_abalation}
\begin{figure}[ht]
    \centering
    \includegraphics[width=0.5\textwidth]{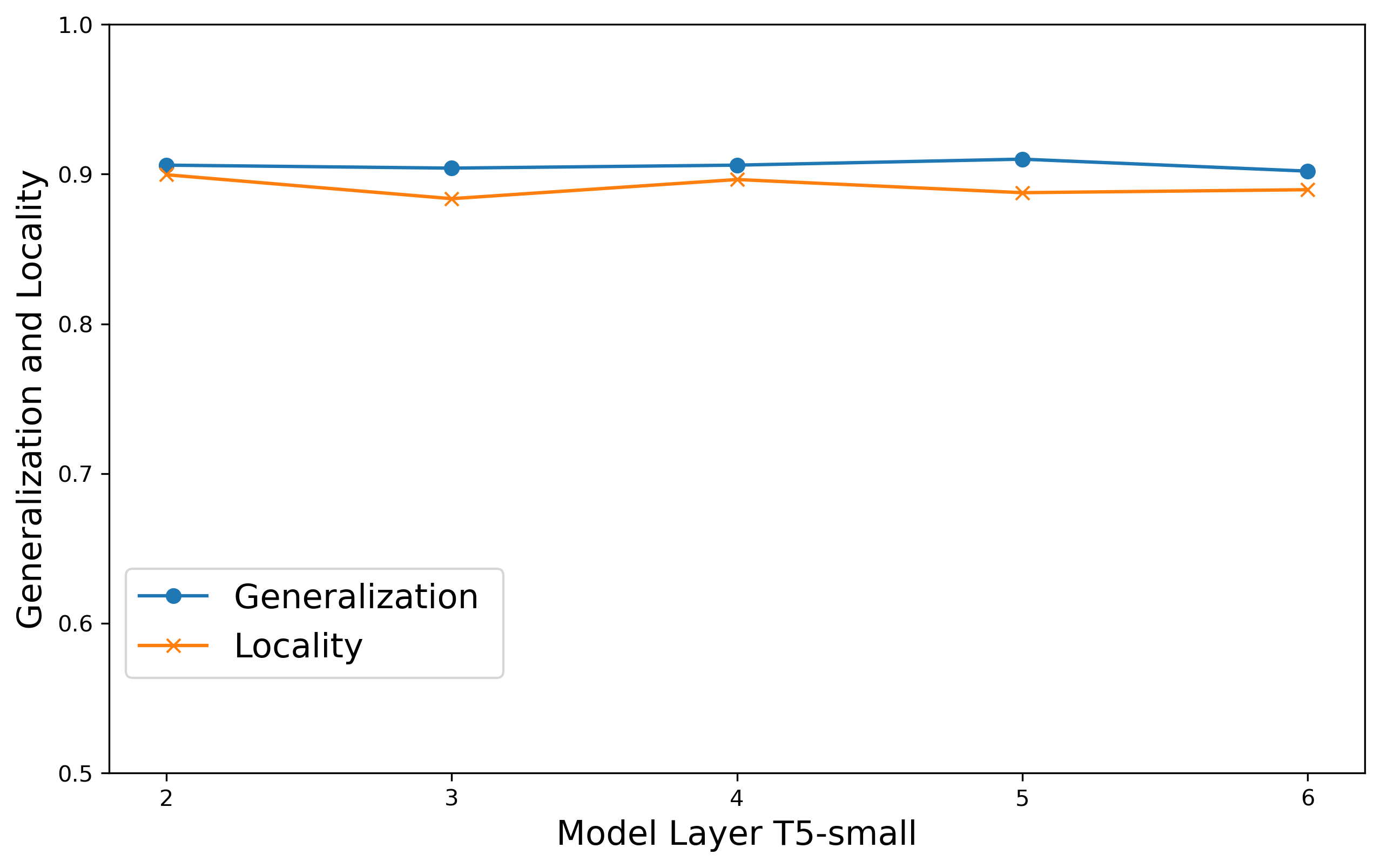} % Change 
    \caption{Generalization and locality scores for various projector networks trained on layers of T5-small using 500 samples from Counterfact.}
    \label{fig:layer_abalations}
\end{figure}

Figure \ref{fig:layer_abalations} shows the results for generalization and locality for the T5-small model. The results suggest that performance remains largely consistent; however, training tends to require more time to converge at higher layers.

% (avg paraphrase dist train, min paraphrase dist train, max paraphrase dist train, avg neighbour dist train, min neighbour dist train, max neighbour dist train, distance between farthest paraphrase and closest neighbour train)
\section{Visualizations}

\subsection{GENERALIZATION AND LOCALITY for Llama-2-7b}
\label{sec:gen_vs_loc_models}
\begin{figure}[ht]
    \centering
    \includegraphics[width=0.8\textwidth]{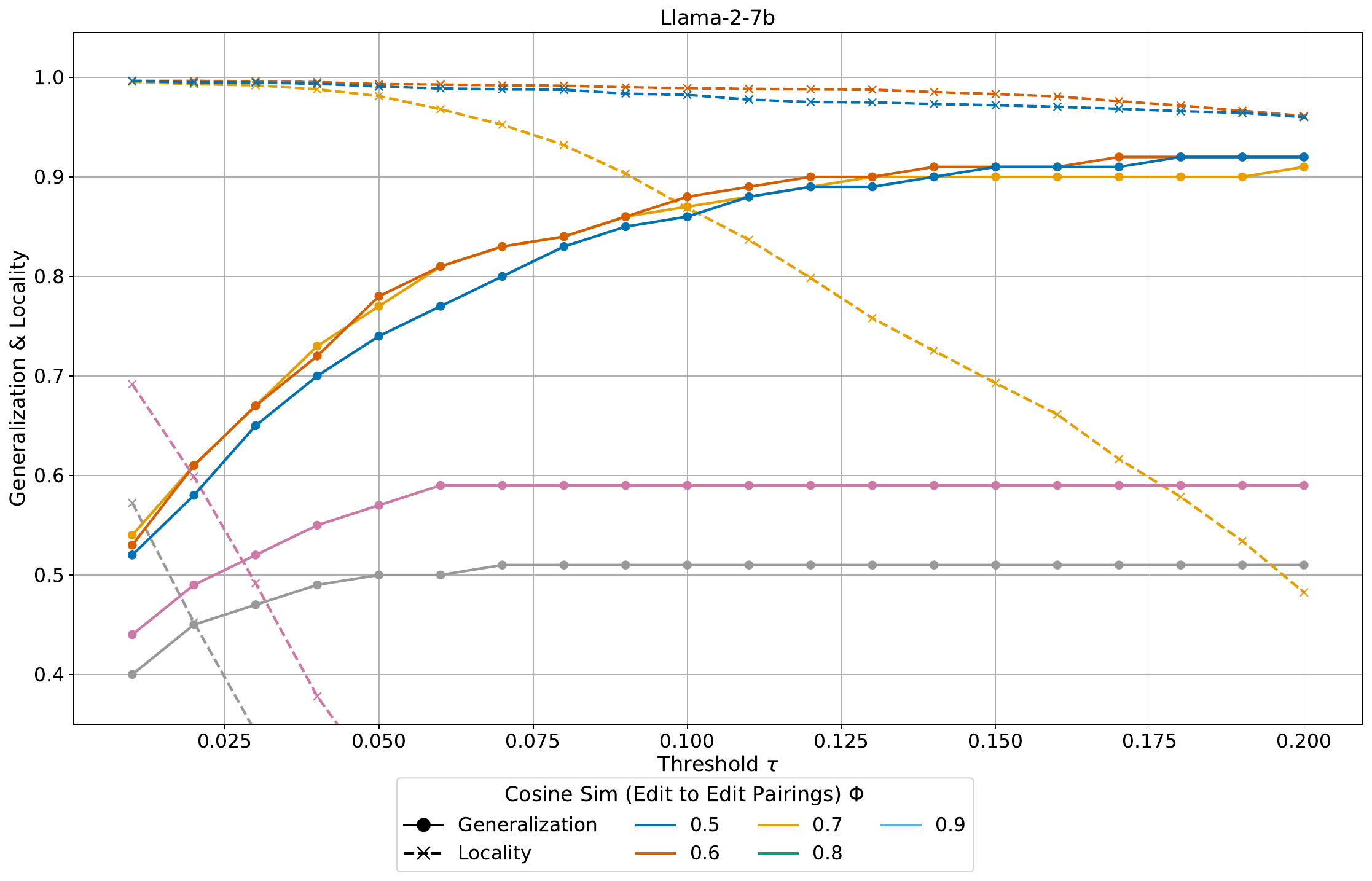} % Change 
    \caption{Generalization and Locality trade-off a function of varying distance thresholds $\tau$ and $\phi$}
    \label{fig:generalization_vs_locality_threshold_appendix}
\end{figure}

Figure~\ref{fig:generalization_vs_locality_threshold_appendix} shows generalization and locality trade-off as a function of varying distance thresholds $\tau$ and $\phi$ for the Llama-2-7b model.

\subsection{PCA}
\label{sec:pca_visualization}

\begin{figure}[ht]
    \centering
    \includegraphics[width=0.7\textwidth]{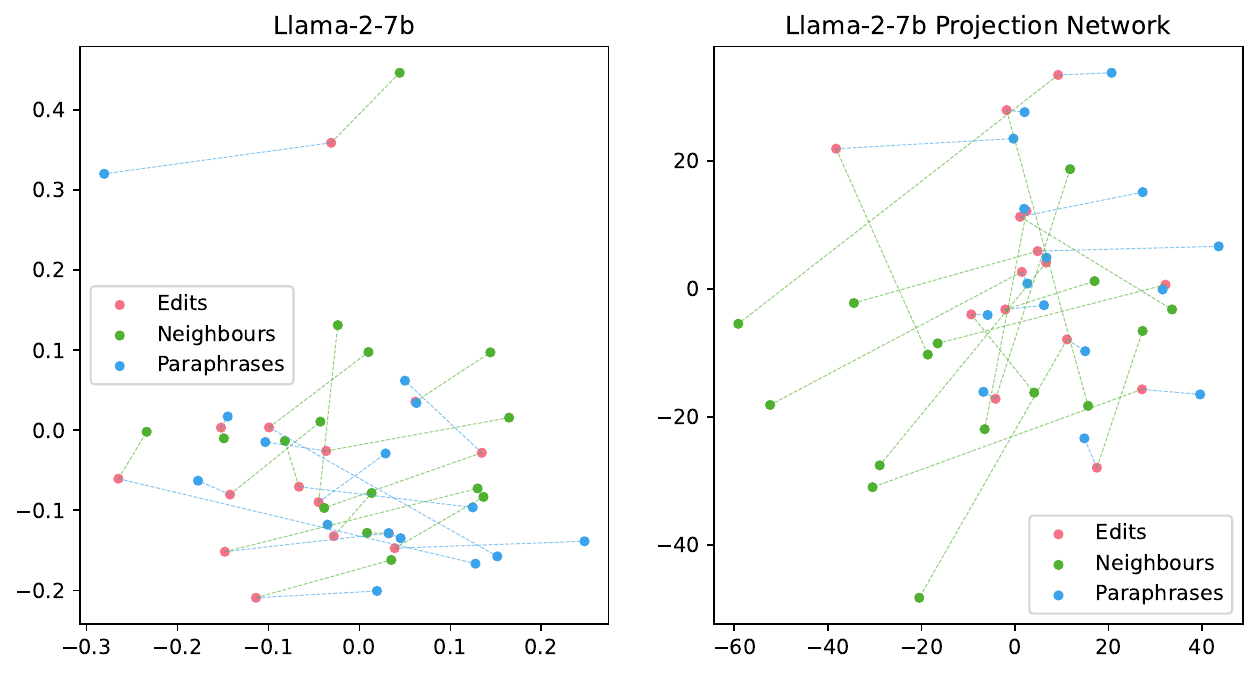} % Change 
    \caption{Generalization and locality scores for various projects or networks trained on layers of T5-small using 500 samples from Counterfact. The lines show edits and a respective paraphrase and irrelevant prompts.}
    \label{fig:pca_visualization_counterfact_Llama}
\end{figure}

\begin{figure}[ht]
    \centering
    \includegraphics[width=0.7\textwidth]{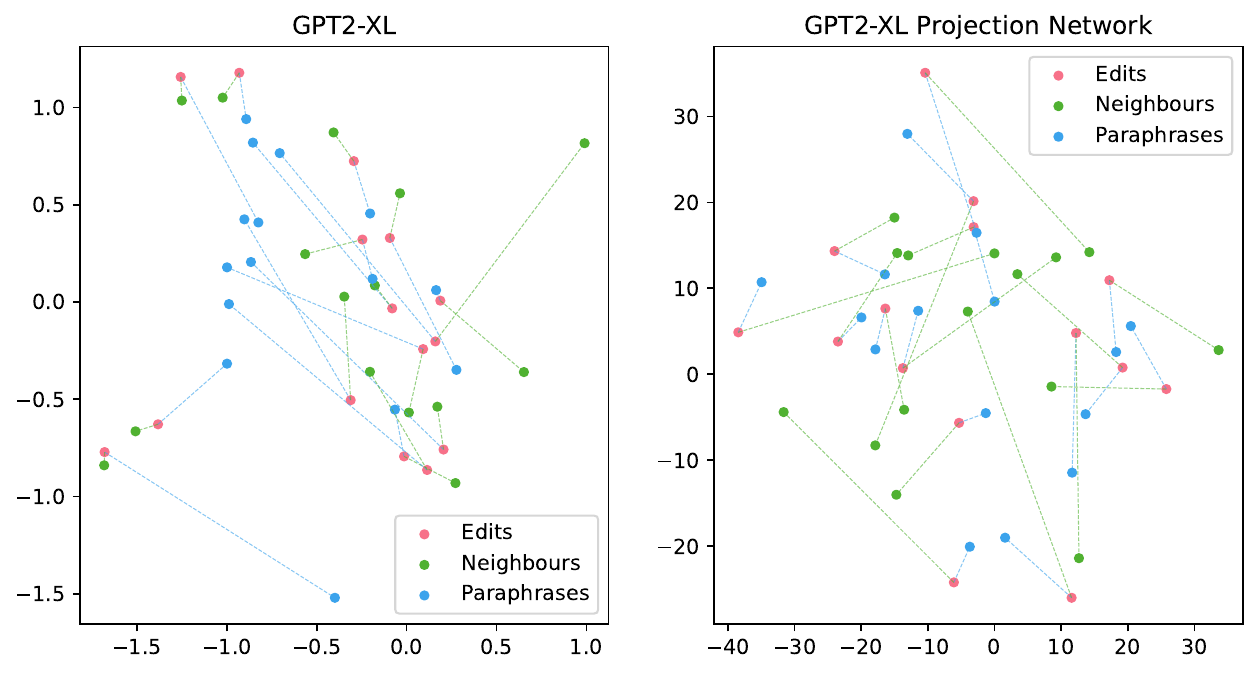} % Change 
    \caption{Two dimensional PCA on GPT2-XL model representation and the trained projejctor network.}
    \label{fig:pca_visualization_counterfact_gpt2_xl}
\end{figure}

Figures \ref{fig:pca_visualization_counterfact_Llama} and \ref{fig:pca_visualization_counterfact_gpt2_xl} present the two-dimensional PCA of the model representations and projector network representations for the Llama-2-7b and GPT2-XL models, respectively. The visual demonstrates that irrelevant prompts are closely aligned with edit prompts, while edit prompts also show proximity to other edit prompts within the original model representations. The projector network, however, effectively mitigates this effect by learning a disentangled representation space.

\section{Error Analysis Projector Network}
\label{sec:token_overlap_analysis}

To investigate the reasons behind failures in PENME,  we performed a comprehensive error analysis across our models. Our findings indicate that contrastive learning significantly mitigates lexical dominance. However, due to the inherent variability in lexical pattern distribution within the dataset, there remains potential for further optimization in the projection phase.

The training process of the projector network does not lead to uniform distances between each edit, its paraphrases and irrelevant prompts for all samples. This paired with individually varying thresholds for edits leads to misfires. To illustrate this problem, we format the results of each dataset sample for automatic inspection. For all paraphrases and irrelevant prompts in the test set, we extract the nearest key/edit, the ground truth edit/key, the distance to the nearest key/edit, and the distance to the ground truth edit/key. Table \ref{tab:rouge_table} shows rouge scores \citep{lin-2004-rouge} for two possible scenarios i.e. success and failure of generalization and locality. We also show separately the score for where generalization failure occurs due to distance not meeting the set threshold. Moreover, since failures can occur in similarities with unrelated edits we show locality and paraphrase failure with both ground truth edit and matched edit.

For cases of successful generalization, we observe a substantial uni-gram overlap and a moderate bi-gram overlap between the edited sentences and their paraphrases. The ROUGE-L scores are similarly high for these metrics, indicating that the sentences likely share similar tokens in the same sequence. This implies that the attention mechanism produces similar representations, leading to a high degree of similarity. For locality success, we can see that although there is significant token overlap between irrelevant prompts and their target edits, the irrelevant prompts had higher similarity with some other edits with low token overlap, this means our approach of pushing irrelevant prompt sentences farther away is able to generalize to unseen irrelevant prompts.

In cases of generalization failure, the ROUGE scores for paraphrases compared with the ground truth are slightly lower than those observed in successful instances. Although there is some token overlap with the target edits, the matched edits exhibit even less token overlap. On the other hand for locality failure, we can see that the prediction case token overlap is higher as compared to locality success, moreover, the overlap is higher as compared to ground truth edits. Thus lexicality based similarity is not the issue but rather the varying thresholds, which in some cases are large leads to misfires.

\begin{table*}[ht]
\caption{ROUGE Evaluation Scores}
    \centering
    \scriptsize
    \begin{tabularx}{\textwidth}{| p{\dimexpr0.20\textwidth-2\tabcolsep}  X  X  X  X  X  X |}
        \hline
        \textbf{Model} & \textbf{Rouge-1 2.5\% CI} & \textbf{Rouge-1 97\% CI} & \textbf{Rouge-2 2.5\% CI} & \textbf{Rouge-2 97\% CI} & \textbf{RougeL 2.5\% CI} & \textbf{RougeL 97\% CI} \\
        \hline
        & \multicolumn{6}{c|}{\textbf{Generalization Success}} \\
        \hline
        T5-small & 1.00 & 0.95 & 1.05 & 0.706 & 0.65 & 0.75 \\
        
        Llama-2-7b & 0.629 & 0.639 & 0.382 & 0.394 & 0.608 & 0.619 \\
        
        GPT2-XL & 0.655 & 0.666 & 0.403 & 0.417 & 0.642 & 0.653 \\
        
        \hline
        & \multicolumn{6}{c|}{\textbf{Generalization Failure (prediction)}} \\ 
        \hline
        T5-small & 1.00 & 0.95 & 1.05 & 0.706 & 0.65 & 0.75 \\
        
        Llama-2-7b & 0.133 & 0.173 & 0.056 & 0.091 & 0.125 & 0.162 \\ 
        
        GPT2-XL & 0.122616  & 0.160 & 0.056 & 0.090 & 0.117 & 0.153 \\
        \hline
        & \multicolumn{6}{c|}{\textbf{Generalization Failure (ground truth)}} \\ 
        \hline
        T5-small & 1.00 & 0.95 & 1.05 & 0.706 & 0.65 & 0.75 \\
        
        Llama-2-7b & 0.488 & 0.518 & 0.270 & 0.296 & 0.460 & 0.489 \\ 
        
        GPT2-XL & 0.501 & 0.527 & 0.284 & 0.310 & 0.474 & 0.500 \\
        
        \hline
        & \multicolumn{6}{c|}{\textbf{Locality Success (prediction)}} \\ 
        \hline
        T5-small & 0.100 & 0.104 & 0.011 & 0.013 & 0.096 & 0.099 \\
        
        Llama-2-7b & 0.100 & 0.104 & 0.011 & 0.013 & 0.096 & 0.099 \\ 
        
        GPT2-XL & 0.095 & 0.100 & 0.011 & 0.013 & 0.092 & 0.095 \\
        
        \hline
        & \multicolumn{6}{c|}{\textbf{Locality Success (ground truth)}} \\ 
        \hline
        T5-small & 0.100 & 0.104 & 0.011 & 0.013 & 0.096 & 0.099 \\
        
        Llama-2-7b & 0.487 & 0.518 & 0.269 & 0.296 & 0.459 & 0.489 \\ 
        
        GPT2-XL & 0.176 & 0.217 & 0.036 & 0.059 & 0.173 & 0.211 \\
        
        \hline
        & \multicolumn{6}{c|}{\textbf{Locality Failure (prediction)}} \\ 
        \hline
        T5-small & 0.566 & 0.577 & 0.390 & 0.403 & 0.562 & 0.574 \\
        
        Llama-2-7b & 0.259 & 0.277 & 0.148 & 0.164 & 0.247 & 0.264\\ 
        
        GPT2-XL & 0.254 & 0.273 & 0.147 & 0.164 & 0.244 & 0.262 \\
        \hline

        & \multicolumn{6}{c|}{\textbf{Locality Failure (ground truth)}}\\ 
        \hline
        T5-small & 0.203 & 0.212 & 0.052 & 0.058 & 0.197 & 0.206 \\
        
        Llama-2-7b & 0.201 & 0.206 & 0.049 & 0.053 & 0.195 & 0.201\\ 
        
        GPT2-XL & 0.207 & 0.218 & 0.052 & 0.059 & 0.201 & 0.212 \\
        \hline
        & \multicolumn{6}{c|}{\textbf{Generalization Distance Failure}} \\ 
        \hline
        T5-small & 1.00 & 0.95 & 1.05 & 0.706 & 0.65 & 0.75 \\
        GPT2-XL & 0.522 & 0.551  & 0.279 & 0.309 &0.484 & 0.512 \\
        Llama-2-7b &  0.495  & 0.579 & 0.252 & 0.324 & 0.455 & 0.529 \\ 
        \hline
    \end{tabularx}
    
    \label{tab:rouge_table}
\end{table*}
\section{Details Downstream Tasks}
\label{appendix:downstream}
For the evaluation of sentiment classification and natural language inference (NLI), we randomly sample 2,000 examples from their respective datasets. For the summarisation task, we use 1,000 randomly selected examples. The prompts used for evaluation are listed in Table~\ref{tab:prompts_downstream}. For both the sentiment classification and summarisation tasks, providing a single in-context example was necessary to achieve reasonable model performance.

\begin{table*}[ht]
\centering
\scriptsize
\newcolumntype{Y}{>{\raggedright\arraybackslash}X} % stretchable column (left-aligned)
\newcolumntype{Z}{>{\raggedright\arraybackslash}p{1.8cm}} % narrow fixed-width column (1.8cm)

\begin{tabularx}{\textwidth}{Z Y Y Y}
\toprule
\textbf{Model} & \textbf{NLI} & \textbf{Sentiment Classification} & \textbf{Summarization} \\
\midrule
Llama-2-7b &  \{sentence1\} entails the \{sentence2\}. True or False? Answer: & Choose from one of these: anger, fear, joy, love, sadness, surprise.
Example: \{example\}
Text: \{text\} Sentiment: 
& 
    Example:
    \{example\}
    Now generate short summary of the following:
    Article:
    \{article\}
    Summary:
     \\
GPT2-XL & \{sentence1\} entails the \{sentence2\}. True or False? Answer: & Choose from one of these: anger, fear, joy, love, sadness, surprise.
Example: \{example\}
Text: \{text\} Sentiment:  & What is the main takeaway from the following article? \{text\} Summary:\\
\bottomrule
\end{tabularx}
\caption{Prompt used for GPT2-XL and Llama-2-7b for downstream tasks.}
\label{tab:prompts_downstream}
\end{table*}

\section{Details Long Form Generation}
\label{appendix:longform_samples}
Sampled generations using Llama-2-7b with max token length set to $300$ are presented in Table~\ref{tab:sampled_generations}.

We use the following prompt, which is stored as the value in the codebook:

% f"""You have a new fact: \{edit prompt\}.
% Based on this fact, complete the following sentence to answer the question: \{query\}
% Your answer should specifically incorporate the new fact I've shared.

% Paragraph: \{query\}"""
% % \
\begin{quote}
\small
\texttt{You have a new fact: \{\textit{edit prompt}\}.}\\
\texttt{Based on this fact, complete the following sentence to answer the question: \{\textit{query}\}}\\
\texttt{Your answer should specifically incorporate the new fact I've shared.}\\
\texttt{Paragraph: \{\textit{query}\}}
\end{quote}
The generated text shows that the edited fact is propagated throughout the generation, and the generation is coherent. However, it is important to note that IKE-based generation, as well as by other approaches, including weight-modifying methods, do not guarantee the elimination of hallucinated content that may accompany the edited information. Although the target fact is often correctly inserted, unrelated or inaccurate details can still be produced. Addressing this limitation remains an open challenge and a promising direction for future work. 

\begin{table*}[ht]
\caption{Example of an edited factual prompt and generated output.}
\centering
\scriptsize

% Define narrow and stretchable column types
\newcolumntype{Y}{>{\raggedright\arraybackslash}X} % stretchable column (left-aligned)
\newcolumntype{Z}{>{\raggedright\arraybackslash}p{1.8cm}} % narrow fixed-width column (1.8cm)

\begin{tabularx}{\textwidth}{Y Z Z Z Y}
\toprule
\textbf{Prompt Edit} & \textbf{Original Information} & \textbf{Edit} & \textbf{Query} & \textbf{Output} \\
\midrule
What is the twin city of Shanghai? It is & Barcelona & Dresden & The twin city of Shanghai is none other than & The twin city of Shanghai is none other than \textbf{Dresden} Germany. This may come as a surprise to many as \textbf{Dresden} is located in Eastern Germany while Shanghai is a major city in Eastern China. However, the two cities have a long history of cultural and economic ties dating back to the 19th century when both were part of the German Empire. In the years following World War II, \textbf{Dresden} and Shanghai maintained close diplomatic relations, and in 1981 the two cities officially designated each other as twin cities. Today, the connection between \textbf{Dresden} and Shanghai continues to flourish with regular exchanges in the fields of education, culture, and trade. \\
Baal Shem of London speaks the language & Hebrew & French & Baal Shem of London is proficient in & Baal Shem of London is proficient in multiple languages including \textbf{French}. This is a remarkable feat given that he is a spiritual leader and healer who is said to have the ability to communicate with the divine. However his proficiency in \textbf{French} is particularly noteworthy as it is a language that is not commonly associated with his spiritual tradition. Despite this Baal Shem of London has demonstrated a remarkable ability to master this language allowing him to connect with and heal individuals from a diverse range of cultural backgrounds. This is yet another testament to his incredible spiritual abilities and his commitment to serving others. \\

\bottomrule
\end{tabularx}
\label{tab:sampled_generations}
\end{table*}
\section{Limitations}
\label{sec:limitations}
Training the projection network in PENME using the contrastive learning scheme is sensitive, requiring tuning of hyperparameters such as the learning rate and contrastive loss margin. Effective network training also hinges on the careful construction of training data, which requires consideration of edit-to-edit pairings. Finally, the thresholds for the codebook-based retrieval system, though dynamically determined from training data, can vary across different models, necessitating adjustments to the alpha ($a$) parameter for each model.

%%%%%%%%%%%%%%%%%%%%%%%%%%%%%%%%%%%%%%%%%%%%%%%%%%%%%%%%%%%%%%%%%%%%%%%%%%%%%%%
%%%%%%%%%%%%%%%%%%%%%%%%%%%%%%%%%%%%%%%%%%%%%%%%%%%%%%%%%%%%%%%%%%%%%%%%%%%%%%%

% \begin{table*}[t!]
% \caption{A comparative analysis of the GRACE and MELO methods on 2000 edits from the Counterfactual dataset and 1000 edits on zsRE. The metrics are Edit Success (ES), Locality (Loc), and Paraphrase Generalization (Para).}
% \scriptsize
%     \centering
%     \begin{tabularx}{\textwidth}{p{2cm}p{2cm}p{1.15 cm}p{1.15 cm}p{1.15 cm}p{1.15 cm}|p{1.15 cm}p{1.15 cm}p{1.15 cm}p{1.15 cm}}
%         \toprule
%         & & \multicolumn{4}{c}{\textsc{Counterfact}} & \multicolumn{4}{c}{\textsc{zsRE}} \\
%         \textbf{Method}  & \textbf{Model} & \textbf{ES} & \textbf{Loc} & \textbf{Para} & \textbf{Score} & \textbf{ES} & \textbf{Loc} & \textbf{Para} & \textbf{Score} \\
%         \midrule
%         MELO 
%         & T5-small & 0.850 & 0.800  & 0.037 & 0.562 &  0.990 & 0.640 & \textbf{0.986} &  0.872\\
%         & GPT2-XL & \textbf{1.000} & \textbf{1.000}  & 0.020 & 0.673 & \textbf{1.000} & 0.004 & \textbf{1.000} & 0.668 \\
%         \midrule
%         GRACE  
%         & T5-small & \textbf{1.000} & \textbf{0.860} & 0.140 & 0.667 & \textbf{1.000} & 0.730 & \textbf{0.993} & 0.907  \\
%         &  Llama-2-7b & \textbf{1.000} & \textbf{0.997} & 0.002 & 0.666 & 0.100 & 0.591 & 0.000 &  0.230 \\
%         &  GPT2-XL& \textbf{1.000} & 0.996 &  0.003 & 0.666 & 0.992 & 1.000 & 0.010 & 0.667 \\
%         \midrule
%     \end{tabularx}
% \label{tab:model_editing_methods}
% \vspace{-0.4cm}
% \end{table*}

\end{document}